\documentclass{article} % For LaTeX2e
\usepackage{iclr2027_conference,times}

\usepackage[utf8]{inputenc}
\usepackage[T1]{fontenc}
\usepackage{hyperref}
\usepackage{url}
\usepackage{caption}
\usepackage{makecell}
\usepackage{booktabs}
\usepackage{mathtools}
\usepackage{amsfonts}
\usepackage{amsmath}
\usepackage{amsthm}
\usepackage{amssymb}
\usepackage{nicefrac}
\usepackage{microtype}
\usepackage{xcolor}
\usepackage{algorithm}
\usepackage{float}
\usepackage{algorithmic}
\usepackage{graphicx}
\usepackage{multirow}
\usepackage{wrapfig}
\usepackage[most]{tcolorbox}
\usepackage{listings}
\usepackage{pifont}
\newcommand{\cmark}{\ding{51}}
\newcommand{\xmark}{\ding{55}}

\definecolor{promptboxblue}{RGB}{210,228,248}
\definecolor{promptboxbluedark}{RGB}{60,110,170}
\newtcolorbox[auto counter, number within=section]{promptbox}[1][]{%
  enhanced,
  colback=promptboxblue!40,
  colbacktitle=promptboxblue!75,
  colframe=promptboxbluedark!85,
  coltitle=black,
  fonttitle=\bfseries\small,
  arc=1.2mm, boxrule=0.4pt,
  left=5pt, right=5pt, top=5pt, bottom=5pt,
  title={#1},
  before skip=8pt, after skip=8pt,
}

\lstdefinestyle{promptlistingstyle}{
  basicstyle=\ttfamily\scriptsize,
  backgroundcolor=\color{promptboxblue!30},
  frame=single,
  rulecolor=\color{promptboxbluedark!85},
  framesep=5pt,
  framerule=0.4pt,
  breaklines=true,
  breakatwhitespace=false,
  columns=fullflexible,
  keepspaces=true,
  showstringspaces=false,
  xleftmargin=5pt,
  xrightmargin=5pt,
  aboveskip=4pt,
  belowskip=4pt,
}

\newcommand{\promptlistingtitle}[1]{%
  \par\smallskip\noindent
  {\setlength{\fboxsep}{2pt}% default is usually 3pt
  \colorbox{promptboxblue!75}{%
    \makebox[0.98\linewidth][l]{\strut\hspace{3pt}\bfseries\small #1\hfill}%
  }}%
  \nopagebreak[4]\par\nobreak
}

\definecolor{failboxred}{RGB}{252,228,228}
\definecolor{failboxreddark}{RGB}{170,55,55}
\lstdefinestyle{failurelistingstyle}{
  basicstyle=\ttfamily\scriptsize,
  backgroundcolor=\color{failboxred!85},
  frame=single,
  rulecolor=\color{failboxreddark!85},
  framesep=5pt,
  framerule=0.4pt,
  breaklines=true,
  breakatwhitespace=false,
  columns=fullflexible,
  keepspaces=true,
  showstringspaces=false,
  xleftmargin=5pt,
  xrightmargin=5pt,
  aboveskip=4pt,
  belowskip=4pt,
}
\newcommand{\faillistingtitle}[1]{%
  \par\smallskip\noindent
  \colorbox{failboxreddark!75}{%
    \makebox[0.98\linewidth][l]{\strut\hspace{4pt}\color{white}\bfseries\small #1\hfill}%
  }%
  \nopagebreak[4]\par\nobreak
}

\newcommand{\pmc}[2]{\mbox{$#1{\scriptstyle\,\pm\,#2}$}}

\title{Internalizing Curriculum Judgment for \\ LLM Reinforcement Fine-Tuning}

\iclrfinalcopy

\author{
Han Zheng$^1$,
Yining Ma$^{1,}$\thanks{Corresponding author: Yining Ma (yiningma@mit.edu).}~,
Karthick Gunasekaran$^2$,
Bharathan Balaji$^2$,\\
Zheng Du$^2$,
Shiv Vitaladevuni$^2$,
and Cathy Wu$^1$\\\\
$^1$MIT\\
$^2$Amazon AGI\\
\texttt{\{hanzheng, yiningma, cathywu\}@mit.edu}\\
\texttt{\{gunaskar, bhabalaj, zhengdu, shivnaga\}@amazon.com}
}

\fancypagestyle{arxivstyle}{
  \fancyhf{}
  \fancyfoot[C]{\thepage}
  \renewcommand{\headrulewidth}{0.4pt}
  \renewcommand{\footrulewidth}{0pt}
}

\begin{document}

\maketitle
\pagestyle{arxivstyle}
\thispagestyle{arxivstyle}

\begin{abstract}
In LLM Reinforcement Fine-Tuning (RFT), curriculum learning drives both efficiency and performance. Yet, current methods externalize curriculum judgment via handcrafted heuristics or auxiliary models, risking misalignment with the policy's training dynamics. In this paper, we introduce METIS (METacognitive Internalized Self-judgment), a novel framework that internalizes curriculum judgment as a native capability. Leveraging a critical observation that within-prompt reward variance effectively gauges prompt informativeness, METIS predicts this metric based on recent training outcomes as lightweight in-context learning examples. This intrinsic self-judgment then dynamically dictates the training allocation. Moreover, METIS closes the loop between judgment and optimization by jointly optimizing the standard RFT rewards and a self-judgment reward. This allows the policy to learn what to learn next, as a form of metacognition. Across mathematical reasoning, code generation, and agentic function-calling benchmarks, METIS delivers superior performance while achieving up to a \(2.1\times\) training speedup, with controlled ablations and in-depth analysis further validating the benefits of internalized curriculum judgment. By bypassing handcrafted heuristics and auxiliary models, our work establishes a simple, closed-loop, and highly efficient curriculum internalization paradigm for LLM reinforcement fine-tuning.
 \end{abstract}

%% ============================================================================
%% ============================================================================
%% ============================================================================
\section{Introduction}
\label{sec:introduction}
%% ============================================================================

Effective learning depends not only on how a learner updates from feedback, but also on what it chooses to practice. Problems already mastered offer little information, while problems far beyond current ability often provide little actionable signal. The most useful practice lies near the boundary of current competence, echoing \emph{zone of proximal development} in education~\citep{vygotsky1978mind}. Because this frontier shifts as learning progresses, a capable learner should not only learn to solve problems, but also monitor their own learning state and decide which problems to learn next. This suggests a form of metacognition: using self-assessment to guide future learning effectively~\citep{Flavell1979MetacognitionAC}.

This principle has become increasingly relevant in reinforcement fine-tuning (RFT) for large language models (LLMs)~\citep{ouyang2022training, shao2024deepseekmath, guo2025deepseek}, which is operationalized as \emph{curriculum learning}. As rollouts are expensive in LLM RFT, the curriculum is not merely a data reordering: it determines how training compute is allocated and therefore directly affects both efficiency and performance. A good curriculum should prioritize prompts that are \emph{informative} for the current policy, avoiding those that are already solved or consistently beyond its capability.

A key signal of such \emph{prompt informativeness} arises naturally in group-relative RFT~\citep{shao2024deepseekmath, yu2026dapo, zheng2025groupsequencepolicyoptimization}, which evaluates multiple rollouts per prompt. When all rollouts receive the same reward, the prompt provides little relative signal since it is consistently solved or failed. In contrast, diverse rollout outcomes indicate that the prompt lies near the policy's competence frontier and can induce stronger updates. Therefore, \emph{within-prompt reward variance} provides a simple proxy for prompt informativeness~\citep{jiang2025vcrl, sun2026improving, gao2026prompt}. In this paper, we formalize and extend such a proxy to handle both discrete and continuous rewards in LLM RFT.

\begin{figure}[!t]
    \centering
    \includegraphics[width=0.70\linewidth]{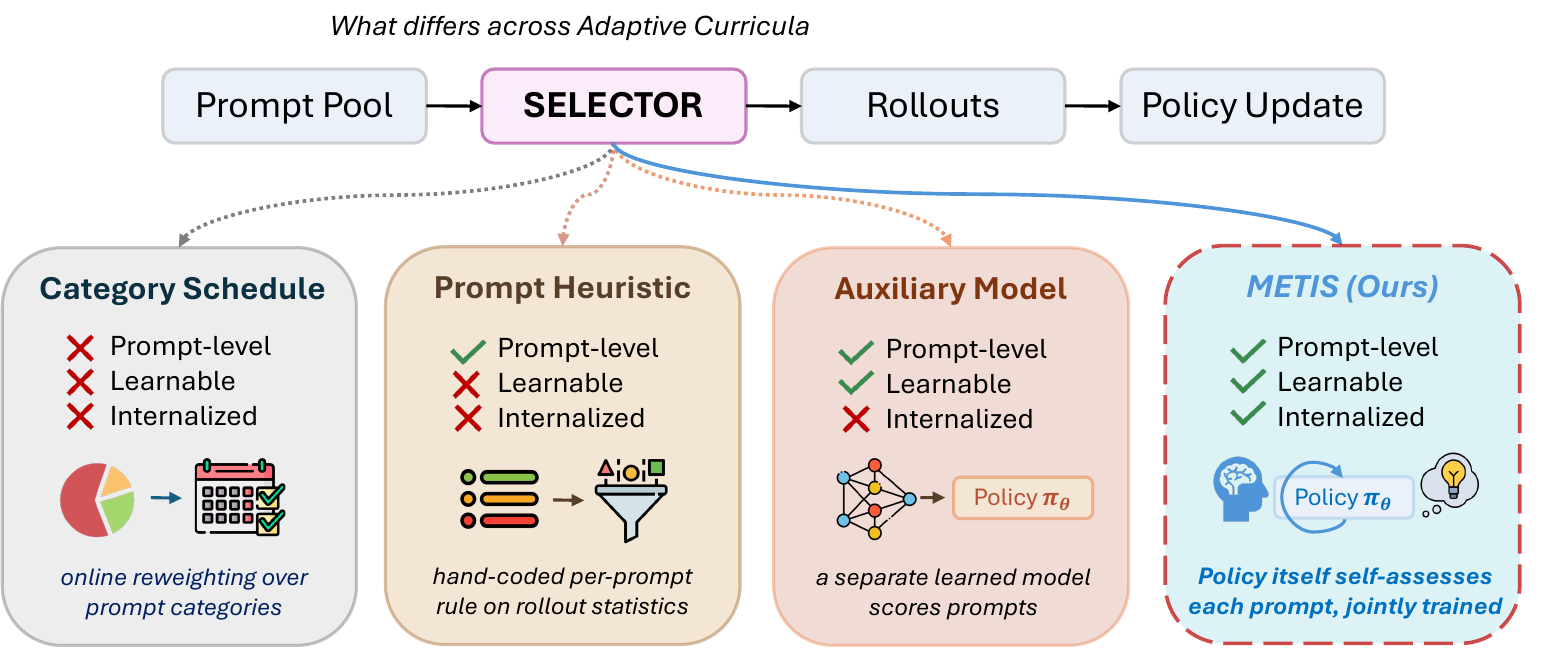}
    \caption{Conceptual comparison. While existing curricula rely on external schedules, heuristics, or auxiliary models, \textsc{METIS} internalizes curriculum judgment via in-context self-assessment, closing the loop by jointly optimizing standard RFT and self-judgment rewards.}
    % \caption{Conceptual comparison of adaptive curriculum mechanisms for LLM RFT. Existing methods produce curriculum judgment through category-level schedules, prompt-level heuristics, or auxiliary selector models - all external to the policy. \textsc{METIS} instead internalizes curriculum judgment within the policy itself: the policy self-assesses candidate prompts via in-context learning, selects prompts predicted to be informative, and improves this judgment from realized rollout outcomes during RFT.}
    \label{fig:conceptual_overview}
    \vspace{-2mm}

\end{figure}

Moreover, prompt informativeness is inherently dynamic. As the policy improves, previously insurmountable prompts may become informative, while once-useful prompts eventually become saturated. Existing curricula address this through various forms of adaptation (see Fig.~\ref{fig:conceptual_overview}): some rely on predefined coarse prompt groups based on domains or data sources~\citep{wang2025dump, chen2025self}, others apply hand-coded heuristics over rollout statistics or offline difficulty labels~\citep{zhang2025learning, parashar2026curriculum, pang2025reasoning, jiang2025vcrl, zheng2026act}, and the rest train a separate model to predict~\citep{sun2026improving, gao2026prompt}. Crucially, curriculum judgment remains \emph{external} to the policy across all these paradigms. But because the competence frontier moves with the policy as it learns, accurate curriculum judgment requires tracking the policy's current state. Thus, we argue that the policy itself is the entity most directly aware of that state. As modern LLMs have strong reasoning capabilities, it could be elicited to read its own rollout outcomes and predict which prompts will be informative. However, current methods neglect such metacognitive potential, treating the policy as a passive recipient rather than an active architect of its own curricula.

% \vspace{-3mm}

This motivates \emph{curriculum internalization}: making curriculum judgment a capability of the policy itself, rather than relying on another auxiliary head. During training, rollout outcomes provide direct evidence about the policy's current learning state. Internalizing curriculum judgment exposes the policy to this evidence directly and trains it to map recent prompt-outcome patterns to future prompt informativeness. The goal is therefore not only to select better prompts for the model, but to make the ability to judge useful prompts part of the model's learned competence. This reframes curriculum learning from selecting data to training the model to decide what it should learn from next.

To this end, we introduce \textsc{METIS} (\textbf{MET}acognitive \textbf{I}nternalized \textbf{S}elf-judgment), a novel framework that internalizes curriculum judgment as
a native policy capability. \textsc{METIS} trains the policy to predict the expected within-prompt reward variance of each prompt candidate as a measure of informativeness, using lightweight in-context examples constructed from recent training outcomes. The predicted scores then determine how the rollout budget is allocated. After rollout, 
the obtained data can be utilized for two coupled learning signals. The original task rewards improve the policy's problem-solving behavior, while we introduce a judgment reward that calibrates the policy's understanding of which prompts are informative. Together, \textsc{METIS} enables the policy to learn not only how to solve tasks, but also what to learn next, closing the loop between judgment and optimization.

Our main contributions are threefold: 1) Conceptually, we formulate curriculum learning for LLM RFT as policy-dependent competence-frontier tracking and identify within-prompt reward variance as a general informativeness signal across discrete and continuous rewards;
2) Technically, we propose \textsc{METIS}, the first LLM RFT framework to internalize curriculum judgment, while jointly optimizing task rewards and a self-judgment reward, transforming curriculum into a learned capability of the policy rather than an external selection mechanism;
3) Empirically, \textsc{METIS} improves performance across mathematical reasoning, code generation, and agentic function-calling benchmarks, yielding up to 2.1$~\times$ wall-clock speedups with only around 3.9\% per-step overhead; controlled ablations and analyses further show that internalized judgment drives these gains while improving prompt coverage and training stability. METIS presents a metacognitive framework for efficient LLM RFT.

\section{Related Work}
\label{sec:related_work}
%% ============================================================================

\textbf{Reinforcement fine-tuning (RFT).} Reinforcement fine-tuning methods vary in reward type and optimization objective. RLHF typically uses preference-based rewards with PPO-style updates~\citep{ouyang2022training, bai2022constitutional, zheng2023ppo}, while critic-free alternatives such as RLOO, ReMax, and REINFORCE++ simplify optimization~\citep{ahmadian2024back, li2023remax, hu2025reinforcestabilizingcriticfreepolicy}. For reasoning tasks, recent work increasingly uses verifiable rewards~\citep{shao2024deepseekmath, guo2025deepseek}, with group-relative methods such as GRPO, DAPO, and GSPO estimating advantages from multiple rollouts per prompt~\citep{shao2024deepseekmath, yu2026dapo, zheng2025groupsequencepolicyoptimization}. 
Orthogonal to the update algorithm, \textsc{METIS} focuses on curriculum learning, i.e., which prompts the policy trains on in each RFT update.

% \textsc{METIS} targets the group-relative family, since it relies on multiple rollouts per prompt to estimate within-prompt reward variance, and is orthogonal to the specific choice within this family. It changes \emph{which} prompts the policy is trained on rather than how the policy is updated.

\textbf{Curriculum learning for RFT.}
We group existing methods by how curriculum judgment is produced (Fig.~\ref{fig:conceptual_overview}). Some methods operate at the category or data-source level, such as DUMP~\citep{wang2025dump} and SEC~\citep{chen2025self}, while others make prompt-level decisions using difficulty estimates, reward dynamics, or rollout statistics, including ADCL~\citep{zhang2025learning}, SPEED-RL~\citep{zhang2025speed}, Dynamic Sampling~\citep{yu2026dapo}, VCRL~\citep{jiang2025vcrl}, and GRESO~\citep{zheng2026act}. More recent methods such as PCL~\citep{gao2026prompt} and attention-based predictors~\citep{sun2026improving} learn the selector itself, but keep it separate from the task policy. In contrast, \textsc{METIS} \emph{internalizes} curriculum judgment by letting the policy make prompt-level curriculum judgments directly and updates this capability online from realized rollout outcomes.

\textbf{In-context learning and model self-assessment.}
LLMs can adapt to new tasks through in-context learning (ICL)~\citep{brown2020language, dong2024survey, wu2025why, liu2022makes, min2022rethinking, rubin2022retrieve} and estimate uncertainty in their own predictions~\citep{kadavath2022language, lin2022teaching, kumaran2026verbalconfidence, ling2024uncertainty, binder2025looking}. \textsc{METIS} unites and extends these two capabilities for curriculum self-judgment in RFT.
% making curriculum self-judgment a trained capability of the policy rather than a fixed elicitation behavior
% curriculum learning. The policy conditions on a calibration memory of recent prompt-variance pairs to predict the rollout variance of candidate prompts before rollout, and this prediction is reinforced online from realized outcomes, making curriculum self-judgment a trained capability of the policy rather than a fixed elicitation behavior.

\section{Problem Formulation}
\label{sec:preliminaries}

We consider reinforcement fine-tuning (RFT) of a language-model policy \(\pi_\theta\) over a prompt dataset \(\mathcal{P}\). For each prompt \(x \in \mathcal{P}\), the policy samples a rollout \(y \sim \pi_\theta(\cdot \mid x)\) and receives a bounded scalar reward \(r(x,y) \in [0,1]\). This formulation covers both binary verifiable rewards (\(r \in \{0,1\}\)), common in mathematical reasoning, and continuous rewards such as partial test-case credit in code generation. Notably, \textsc{METIS} operates uniformly across \emph{both} regimes, whereas pass-rate heuristics used in prior work~\citep{yu2026dapo, jiang2025vcrl, sun2026improving, gao2026prompt, chen2025self, wang2025dump, zhang2025learning} are restricted to the binary case.

\textbf{Group-relative reinforcement fine-tuning.}
We develop \textsc{METIS} on top of the prevailing group-relative RFT methods that sample multiple rollouts per prompt~\citep{shao2024deepseekmath, yu2026dapo, zheng2025groupsequencepolicyoptimization, minimax2025m1}. At iteration \(t\), a batch of prompts \(\mathcal{S}_t\) is selected for rollout. For each prompt \(x \in \mathcal{S}_t\), the policy generates \(n\) rollouts \(\{y_i\}_{i=1}^{n}\) with rewards \(r_i = r(x,y_i)\), and a centered group-relative advantage is constructed as
\begin{equation}
    A_i(x) = r_i - \bar r(x),
    \qquad
    \bar r(x) = \frac{1}{n}\sum_{j=1}^{n} r_j.
    \label{eq:centered_advantage}
\end{equation}
Following~\cite{liu2025understandingr1zeroliketrainingcritical}, we drop the per-prompt standard-deviation normalization originally used in GRPO~\citep{guo2025deepseek}.The resulting policy loss takes a clipped surrogate form:
\begin{equation}
    \mathcal{L}_{\mathrm{policy}}(\theta)
    =
    -\frac{1}{|\mathcal{S}_t|n}
    \sum_{x \in \mathcal{S}_t}\sum_{i=1}^{n}
    \min\!\left(\rho_i(\theta) A_i(x),\;
        \mathrm{clip}\bigl(\rho_i(\theta),1-\epsilon,1+\epsilon\bigr) A_i(x)\right),
    \label{eq:policy_loss}
\end{equation}
where \(\rho_i(\theta) = \pi_\theta(y_i \mid x) / \pi_{\theta_{\mathrm{old}}}(y_i \mid x)\). Note that \textsc{METIS} is agnostic to the specific normalization, clipping, or KL-regularization choices made by individual algorithms. It requires only that multiple rollout rewards are available per prompt so that within-prompt outcome variability can be estimated.

\textbf{Quantifying prompt informativeness via within-prompt reward variance.}
In group-relative RFT, prompts with diverse rollout rewards yield stronger relative learning signals than those with similar rewards~\citep{gao2026prompt}. We formalize this and extend it to continuous rewards: by Eq.~\eqref{eq:centered_advantage}, the mean squared advantage for prompt $x$ equals the empirical reward variance across its $n$ rollouts,
\begin{equation}
    v(x) \;\coloneqq\; \frac{1}{n}\sum_{i=1}^{n} A_i(x)^2 \;=\; \frac{1}{n}\sum_{i=1}^{n}\bigl(r_i - \bar r(x)\bigr)^2,
    \label{eq:reward_variance}
\end{equation}
which vanishes when all rollouts receive the same reward (uniformly solved or failed prompts) and increases as their outcomes become more diverse. To remove the dependence on a particular set of sampled rollouts, we define the corresponding population quantity under the current policy: \begin{equation} v_\theta(x) \;\coloneqq\; \mathrm{Var}_{y\sim\pi_\theta(\cdot\mid x)} \!\left[r(x,y)\right], \label{eq:expected_variance} \end{equation} with \(v(x)\) serving as its empirical estimate. Importantly, we note that the subscript $\theta$ is essential, as \(v_\theta(x)\) is \emph{policy-dependent} and changes throughout training. For rewards bounded in \([0,1]\), \(v_\theta(x)\leq 1/4\), with equality attained when the reward takes values \(0\) and \(1\) with equal probability.\footnote{For any \(R\in[0,1]\) with mean \(\mu\), \(R^2\leq R\) implies \(\mathrm{Var}(R)\leq\mu(1-\mu)\leq1/4\).}

The same formulation applies to both binary and continuous rewards. For binary rewards, \(v_\theta(x)=p(x)(1-p(x))\), where \(p(x)\) is the per-prompt success probability and estimated empirically by the rollout pass rate. It is maximized at \(p(x)=1/2\), recovering the intermediate-difficulty criterion used in prior binary-curriculum methods~\citep{jiang2025vcrl, zheng2026act, gao2026prompt, sun2026improving}. For continuous rewards, a pass rate is no longer defined, but \(v_\theta(x)\) still captures variation in rollout quality and thus the strength of the learning signal. This allows \textsc{METIS} to handle tasks such as code generation and agentic function calling without a task-specific notion of difficulty.

\textbf{Curriculum as competence-frontier tracking.}
A \emph{curriculum} is a selector that constructs a batch of \(B\) prompts for rollout. Taking informativeness to be the policy-dependent variance \(v_\theta(x)\) in Eq.~\eqref{eq:expected_variance}, the \emph{ideal} curriculum selects the most informative prompts from a candidate pool \(\mathcal{C}_t\):
\begin{equation}
    \mathcal{S}_t^\star
    =
    \mathrm{TopB}_{x\in\mathcal{C}_t} v_\theta(x),
    \qquad
    |\mathcal{S}_t^\star| = B.
    \label{eq:oracle_selection}
\end{equation}
Since \(v_\theta(x)\) changes with the policy, the selected prompts form a moving \emph{competence frontier} throughout training. Existing methods approximate this frontier using external schedules, heuristics, or auxiliary selector models~(e.g., \cite{gao2026prompt}). In each case, informativeness is externally determined rather than learned or refined by the policy, risking misalignment with its evolving competence. \textsc{METIS} instead trains the policy to predict prompt informativeness, as detailed in Sec.~\ref{sec:method}.

\begin{figure}[!t]
    \centering
    \includegraphics[width=\linewidth]{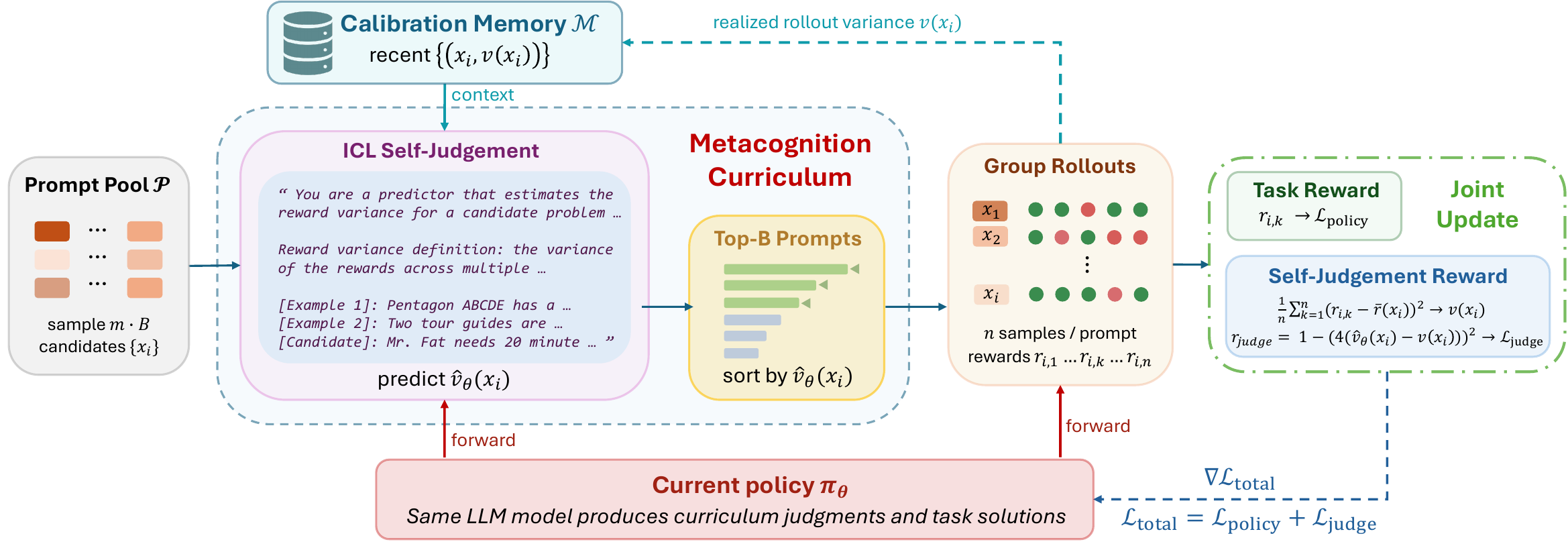}
    % \vspace{-3mm}
    \caption{{Overview of \textsc{METIS}.} At each iteration, the policy predicts candidate informativeness $\hat v_\theta(x)$ via in-context learning on a calibration memory of recent prompt-variance pairs (\textbf{left}). The most informative prompts are then rolled out (\textbf{middle}) to yield task rewards and realized variance $v(x)$. The policy is jointly optimized (\textbf{right}) via standard task loss on solution tokens and a self-judgment loss on prediction tokens, while the calibration memory refreshes with the latest variances.}
    
    \label{fig:method_overview}
    \vspace{-2mm}
\end{figure}
%% ============================================================================
%% ============================================================================
\section{Method}
\label{sec:method}
%% ============================================================================

\textsc{METIS} internalizes curriculum learning as a native metacognitive skill via a two-fold design (see Fig.~\ref{fig:method_overview}). First, the policy self-judges prompt informativeness pre-rollout through \emph{in-context learning} (Sec.~\ref{sec:self_judgment}). Second, the realized rollout variance provides a calibration reward to update this predictive capability online, closing the loop between task optimization and curriculum judgment (Sec.~\ref{sec:internalization}--\ref{sec:joint_optimization}).

\subsection{Self-Judgment via In-Context Learning}
\label{sec:self_judgment}

The previous section framed curriculum selection as predicting the policy-dependent informativeness \(v_\theta(x)\) of each candidate prompt and selecting those with the highest predicted values. In \textsc{METIS}, these predictions are generated by the policy itself through prompt-level in-context self-assessment. Given a candidate \(x\), the policy conditions on a compact context drawn from the recent training history \(\mathcal{H}_t\) and produces a short structured prediction of its expected informativeness.

We use ICL for curriculum self-judgment for four main reasons: \textbf{(i) Predictive.} The policy estimates \(v_\theta(x)\) before task rollouts, enabling candidates to be ranked by predicted informativeness rather than filtered only after observing outcomes~\citep{yu2026dapo, wang2025dump, chen2025self}. \textbf{(ii) Adaptive.} Recent prompt-variance examples reflect the policy's current competence, allowing its judgment to evolve throughout training. \textbf{(iii) Efficient.} Each candidate requires only a short judgment response, rather than the \(n\) full rollouts needed to estimate variance directly~\citep{zhang2025learning, jiang2025vcrl}, adding only a small fraction of the per-step training cost (Sec.~\ref{mainexp}). \textbf{(iv)~Compatible.} Unlike a separate value-head predictor (see more discussions and results in Sec.~\ref{sec:mainablation}), ICL judgments are expressed as language tokens, making them compatible with the original task tokens and facilitating more natural joint optimization and internalization in Sec.~\ref{sec:joint_optimization}.

We instantiate the in-context evidence as a \textbf{calibration memory}
\(
    \mathcal{M}_t
    =
    \{(x_k,v(x_k))\}_{k=1}^{K}
    \subset \mathcal{H}_t,
\)
using prompt-variance pairs from the previous training step, providing recent examples of informativeness under the evolving policy. To include examples with low, medium, and high informativeness as ICL examples, we use three reference variance values, \(0\), \(0.10\), and \(0.25\). For each value, we select the \(K/3\) prompts whose realized variances are closest to it. Note that these reference variances are used only for exemplar retrieval; the stored variances themselves remain their realized values.  

For each candidate \(x\in\mathcal{C}_t\), we then construct
\(
    \mathrm{ctx}_t(x)
    =
    [\,I_{\mathrm{judge}};\mathcal{M}_t;x\,],
\)
where \(I_{\mathrm{judge}}\) specifies the instructions for the variance-prediction task and output format. Since \(v_\theta(x)\in[0,1/4]\) (Sec.~\ref{sec:preliminaries}), the instruction asks the policy to assess the candidate and generate a short response \(z_x\sim\pi_\theta(\cdot\mid\mathrm{ctx}_t(x))\) containing a boxed estimate in this range (e.g., \texttt{\textbackslash boxed\{0.15\}}). We deterministically parse this value as the predicted informativeness, \(\hat{v}_\theta(x)=\mathrm{Parse}(z_x)\). Responses without a valid prediction are treated as judgment failures (App.~\ref{app:failure_boundary}). The rollout batch is then selected as
\(
    \mathcal{S}_t
    =
    \mathrm{TopB}_{x\in\mathcal{C}_t}\hat v_\theta(x).
\)
The full prompt template and a sample judgment response are provided in App.~\ref{app:prompt_template} and App.~\ref{app:judgment_example}.

 This formulation keeps curriculum judgment in the same token-generation space as task solving, leveraging the policy's existing ICL and self-evaluation capabilities~\citep{ren2023selfevaluationimprovesselectivegeneration, kumaran2026verbalconfidence}. Prompt selection is thus driven directly by the policy's own judgment rather than by an external heuristic or auxiliary selector.

\subsection{Internalizing Curriculum Judgment}
\label{sec:internalization}
In-context elicitation alone makes the policy's self-judgment \emph{available}, but does not make it \emph{accurate}.
% where \(\hat v_\theta(x)\) remains a fixed elicitation behavior whose quality depends on whatever calibration ICL happens to provide. 
 \textsc{METIS} closes this gap by reinforcing curriculum judgment from realized outcomes.

For each selected prompt \(x \in \mathcal{S}_t\), we generate \(n\) task rollouts, compute rewards \(\{r_i\}_{i=1}^{n}\), and obtain the realized variance \(v(x)\) from Eq.~\eqref{eq:reward_variance}.  We score each pre-rollout prediction against its realized counterpart with a squared-error calibration reward~\citep{glenn1950verification},
\begin{equation}
    R_{\mathrm{judge}}(x) \;=\; 1 - \bigl(4(\hat v_\theta(x) - v(x))\bigr)^2,
    \label{eq:judge_reward}
\end{equation}

which is maximized only when the pre-rollout judgment matches the realized variance. The supervision is fully endogenous, coming from the rollouts the policy just generated as part of normal RFT, with no \emph{additional labels}, no \emph{auxiliary model}, and no \emph{extra rollout cost}.

Because each prompt produces a single judgment response, we optimize the judgment tokens with a REINFORCE-style objective using an exponential moving-average (EMA) baseline~\citep{ahmadian2024back}:
\begin{equation}
    \nabla_\theta J_{\mathrm{judge}}
    =
    \frac{1}{|\mathcal{S}_t|}
    \sum_{x\in\mathcal{S}_t}
    \bigl(R_{\mathrm{judge}}(x)-b_t\bigr)
    \nabla_\theta
    \log \pi_\theta
    \bigl(z_x \mid \mathrm{ctx}_t(x)\bigr),
    \label{eq:judge_gradient}
\end{equation} where we denote the corresponding minimization objective by \(\mathcal{L}_{\mathrm{judge}}\). The EMA baseline is defined as follows, where \(\alpha\in[0,1)\) controls smoothing rate:
\begin{equation}
    b_{t}
    =
    \alpha b_{t-1}
    +
    (1-\alpha)\,\overline{R}_{\mathrm{judge},t},
    \qquad
    \overline{R}_{\mathrm{judge},t}
    =
    \frac{1}{|\mathcal{S}_t|}
    \sum_{x\in\mathcal{S}_t}
    R_{\mathrm{judge}}(x).
    \label{eq:judge_baseline}
\end{equation}

This objective makes curriculum judgment a trainable behavior of the policy: predictions that correctly anticipate informative prompts are reinforced, while miscalibrated judgments (e.g., overconfident on saturated prompts, or dismissive of high-variance ones) are suppressed. As training progresses, this allows the induced curriculum to tightly track the policy's advancing competence frontier.

\subsection{Joint Optimization}
\label{sec:joint_optimization}

\textsc{METIS} jointly optimizes two token streams with the same policy: task-solution tokens under the group-relative RFT loss \(\mathcal{L}_{\mathrm{policy}}\) in Eq.~\eqref{eq:policy_loss}, and self-judgment tokens under the calibration loss \(\mathcal{L}_{\mathrm{judge}}\):
\begin{equation}
    \mathcal{L}_{\mathrm{total}}
    =
    \mathcal{L}_{\mathrm{policy}}
    +
    \lambda\,\mathcal{L}_{\mathrm{judge}},
    \label{eq:total_loss}
\end{equation}
where \(\lambda\) controls the judgment-update strength. Jointly updating the same policy, task rewards improve problem solving while calibration rewards improve prediction of prompt informativeness, allowing the curriculum to co-evolve with competence without hand-tuned schedules, predefined difficulty, or auxiliary selectors. Algorithm~\ref{alg:metis} summarizes the full procedure.

\begin{algorithm}[t]
\caption{\textsc{METIS}: Metacognitive Internalized Self-Judgment for RFT}
\label{alg:metis}
\begin{algorithmic}[1]
\REQUIRE policy \(\pi_\theta\); prompt dataset \(\mathcal{P}\); batch size \(B\), pool multiplier \(m\), group rollouts size \(n\), judgment weight \(\lambda\), EMA rate \(\alpha\), self-judgment instruction \(I_{\mathrm{judge}}\)
\STATE Initialize baseline \(b \leftarrow 0\)
\FOR{iteration \(t = 1, 2, \ldots\)}
    \STATE Construct the calibration memory \(\mathcal{M}_t\) from training history \(\mathcal{H}_t\).
    \STATE Sample \(\mathcal{C}_t \subset \mathcal{P}\), \(|\mathcal{C}_t|=mB\).
    \STATE For each \(x\in\mathcal{C}_t\): generate \(z_x\sim\pi_\theta(\cdot\mid[I_{\mathrm{judge}};\mathcal{M}_t;x])\) and parse \(\hat v_\theta(x)\).
    \STATE Select \(\mathcal{S}_t \leftarrow \mathrm{TopB}_{x \in \mathcal{C}_t}\, \hat v_\theta(x)\).
    \STATE For each \(x \in \mathcal{S}_t\): sample \(\{y_i\}_{i=1}^n \sim \pi_\theta(\cdot \mid x)\), compute rewards \(\{r_i\}_{i=1}^{n}\), realized variance \(v(x) = \tfrac{1}{n}\sum_i (r_i - \bar r(x))^2\), and \(R_{\mathrm{judge}}(x) = 1-(4(\hat v_\theta(x) - v(x)))^2\).
    \STATE Update \(\theta\) on \(\mathcal{L}_{\mathrm{total}} = \mathcal{L}_{\mathrm{policy}} + \lambda\, \mathcal{L}_{\mathrm{judge}}\).
    \STATE Refresh \(\mathcal{H}_t\) with \(\{(x, v(x))\}_{x \in \mathcal{S}_t}\); update baseline \(b \leftarrow \alpha b + (1-\alpha)\,\overline{R}_{\mathrm{judge}}\).
\ENDFOR
\end{algorithmic}
\end{algorithm}

%% ============================================================================

%% ============================================================================
\section{Experiments}
\label{sec:experiments}

% {\color{red} We depict our experiments in this section, with additional ones in App.~\ref{}. }

\textbf{Models and datasets.}
We span three task families covering discrete and continuous rewards across single and multi-turn rollouts. For \emph{mathematical reasoning} (binary reward), we train on DAPO-17k~\citep{yu2026dapo} and evaluate on AIME24/25~\citep{maa2024aime}, MATH500~\citep{hendrycks2021measuring}, and Minerva~\citep{lewkowycz2022solving}. For \emph{code generation} (continuous reward), we train on CodeContests+~\citep{wang2025codecontestshighqualitytestcase} and evaluate on LiveCodeBench v6~\citep{jain2025livecodebench}, HumanEval~\citep{chen2021evaluating}, MBPP~\citep{austin2021program}, and BigCodeBench~\citep{zhuo2025bigcodebench}. For \emph{agentic function-calling} (continuous, multi-turn), we train and evaluate on BFCL v3~\citep{patil2025bfcl}. We use Qwen3-4B/8B-Base/Instruct~\citep{yang2025qwen3}, Llama-3.1-8B-Instruct~\citep{grattafiori2024llama}, and DeepSeek-R1-Distill-Llama-8B~\citep{guo2025deepseek} as backbone models. Dataset preprocessing and evaluation details are provided in App.~\ref{app:dataset_preprocessing}.

\textbf{RFT backbone and engine.}
All methods use GRPO~\citep{shao2024deepseekmath, guo2025deepseek} running via the VERL framework~\citep{sheng2025hybridflow} on identical hardware. Full details are in App.~\ref{app:training_hparams}.

\textbf{Baselines.} We compare \textsc{METIS} with \textbf{No Curriculum} (uniform prompt sampling) and one representative method from each baseline category in Fig.~\ref{fig:conceptual_overview}: \textbf{SEC}~\citep{chen2025self} for category-level scheduling, \textbf{ADCL}~\citep{zhang2025learning} for heuristic prompt-level curricula, and \textbf{PCL}~\citep{gao2026prompt} for auxiliary-model-based selection. Since these methods were originally developed for binary rewards, we adapt them to continuous-reward tasks where necessary (App.~\ref{app:baselines}). We omit closely related methods within the same categories to avoid redundant comparisons. DUMP~\citep{wang2025dump} is similar in spirit to SEC; VCRL~\citep{jiang2025vcrl} and \citet{sun2026improving} are not readily applicable to continuous rewards; and SPEED-RL~\citep{zhang2025speed}, Dynamic Sampling~\citep{yu2026dapo}, and GRESO~\citep{zheng2026act} have already been shown to be outperformed by PCL.

\begin{table*}[t]
\centering
\caption{
Main results (mean \(\pm\) std over three seeds). \textbf{Avg.} denotes the average Pass@1 at the best checkpoint, \textbf{Time} denotes the hours required to reach it, and \textbf{Spd.} denotes the speedup over No Curriculum, \(T_{\mathrm{No\ Curr.}}/T_{\mathrm{method}}\). \textbf{Bold} and \underline{underlined} values indicate the best and the second-best.
}
\label{tab:main_results}

% \vspace{3mm}
\setlength{\tabcolsep}{2pt}
\renewcommand{\arraystretch}{1.05}
\scriptsize

\resizebox{\textwidth}{!}{%
\begin{tabular}{
    @{}l
    *{7}{c}
    @{\hspace{8pt}}
    *{7}{c}
    @{}
}
\toprule

& \multicolumn{7}{c}{\textbf{DAPO-17K (Mathematical Reasoning)}}
& \multicolumn{7}{c}{\textbf{CodeContests+ (Code Generation)}} \\

\cmidrule(lr){2-8}
\cmidrule(lr){9-15}

\textbf{Method}
& \textbf{AIME24}
& \textbf{AIME25}
& \textbf{MATH500}
& \textbf{Minerva}
& \textbf{Avg.\,$\uparrow$}
& \textbf{Time\,$\downarrow$}
& \textbf{Spd.\,$\uparrow$}
& \textbf{LCBv6}
& \textbf{HumanEval}
& \textbf{MBPP}
& \textbf{BCB}
& \textbf{Avg.\,$\uparrow$}
& \textbf{Time\,$\downarrow$}
& \textbf{Spd.\,$\uparrow$} \\

\midrule

% =========================================================
% Block 1
% =========================================================
\multicolumn{1}{@{}l}{\textbf{Model}}
& \multicolumn{7}{c}{\textbf{Qwen3-8B Base}}
& \multicolumn{7}{c}{\textbf{Qwen3-8B Base}} \\

\cmidrule(lr){2-8}
\cmidrule(lr){9-15}

No Curr.
& \pmc{20.3}{0.9}
& \pmc{17.9}{0.5}
& \pmc{77.7}{1.2}
& \pmc{27.0}{2.2}
& \pmc{35.7}{0.7}
& \pmc{31.4}{3.8}
& $1.0\times$
& \pmc{31.6}{0.5}
& \pmc{85.2}{0.4}
& \pmc{68.9}{0.8}
& \pmc{26.7}{1.1}
& \pmc{53.1}{0.7}
& \pmc{16.7}{5.7}
& $1.0\times$ \\

SEC
& \pmc{20.6}{2.0}
& \pmc{18.3}{0.6}
& \pmc{76.9}{1.6}
& \pmc{27.8}{1.7}
& \pmc{35.9}{0.8}
& \pmc{20.9}{6.6}
& $1.5\times$
& \pmc{32.5}{0.8}
& \pmc{83.5}{1.1}
& \pmc{68.9}{0.2}
& \pmc{27.8}{1.8}
& \pmc{53.2}{0.9}
& \pmc{16.0}{6.4}
& $1.0\times$ \\

ADCL
& \pmc{20.9}{1.1}
& \pmc{18.7}{1.0}
& \pmc{74.9}{1.6}
& \pmc{27.9}{2.2}
& \pmc{35.6}{0.8}
& \pmc{34.6}{5.4}
& $0.9\times$
& \pmc{31.3}{0.3}
& \pmc{85.0}{1.3}
& \pmc{69.1}{0.8}
& \pmc{26.2}{0.9}
& \pmc{52.9}{0.8}
& \pmc{24.3}{6.6}
& $0.7\times$ \\

PCL
& \pmc{21.4}{1.9}
& \pmc{19.6}{0.2}
& \pmc{76.5}{1.6}
& \pmc{27.3}{2.4}
& \pmc{\underline{36.2}}{0.5}
& \pmc{\underline{19.4}}{3.2}
& $\underline{1.6\times}$
& \pmc{31.5}{0.2}
& \pmc{85.6}{1.3}
& \pmc{68.3}{0.8}
& \pmc{27.7}{2.4}
& \pmc{\underline{53.3}}{0.7}
& \pmc{\underline{15.0}}{3.6}
& $\underline{1.1\times}$ \\

\textbf{\textsc{METIS}}
& \pmc{22.1}{1.0}
& \pmc{19.0}{0.3}
& \pmc{77.8}{1.1}
& \pmc{28.2}{0.6}
& \pmc{\textbf{36.8}}{0.4}
& \pmc{\textbf{15.3}}{2.2}
& $\mathbf{2.1\times}$
& \pmc{32.2}{0.3}
& \pmc{86.2}{0.4}
& \pmc{70.2}{0.5}
& \pmc{27.8}{1.0}
& \pmc{\textbf{54.1}}{0.4}
& \pmc{\textbf{11.5}}{2.8}
& $\mathbf{1.5\times}$ \\

\midrule

% =========================================================
% Block 2
% =========================================================
\multicolumn{1}{@{}l}{\textbf{Model}}
& \multicolumn{7}{c}{\textbf{Qwen3-4B Base}}
& \multicolumn{7}{c}{\textbf{Llama-3.1-8B-Instruct}} \\

\cmidrule(lr){2-8}
\cmidrule(lr){9-15}

No Curr.
& \pmc{17.2}{1.0}
& \pmc{10.4}{0.8}
& \pmc{64.1}{2.9}
& \pmc{18.6}{2.1}
& \pmc{27.6}{1.0}
& \pmc{11.9}{2.4}
& $1.0\times$
& \pmc{17.5}{0.4}
& \pmc{64.4}{1.4}
& \pmc{57.7}{0.5}
& \pmc{18.4}{0.6}
& \pmc{39.5}{0.7}
& \pmc{13.2}{6.2}
& $1.0\times$ \\

SEC
& \pmc{15.1}{1.4}
& \pmc{9.4}{1.5}
& \pmc{63.5}{1.9}
& \pmc{17.6}{1.3}
& \pmc{26.4}{0.8}
& \pmc{\textbf{8.4}}{2.1}
& $\mathbf{1.4\times}$
& \pmc{16.4}{1.2}
& \pmc{65.4}{0.7}
& \pmc{58.2}{0.2}
& \pmc{20.0}{0.8}
& \pmc{40.0}{0.7}
& \pmc{14.5}{5.9}
& $0.9\times$ \\

ADCL
& \pmc{17.0}{1.0}
& \pmc{9.2}{0.7}
& \pmc{64.9}{0.7}
& \pmc{16.8}{1.7}
& \pmc{27.0}{0.6}
& \pmc{15.8}{3.1}
& $0.8\times$
& \pmc{17.8}{0.4}
& \pmc{64.4}{1.8}
& \pmc{57.9}{0.6}
& \pmc{18.6}{1.1}
& \pmc{39.7}{0.8}
& \pmc{18.0}{3.8}
& $0.7\times$ \\

PCL
& \pmc{18.4}{1.5}
& \pmc{11.4}{1.1}
& \pmc{65.1}{0.1}
& \pmc{17.8}{0.8}
& \pmc{\underline{28.2}}{0.5}
& \pmc{13.8}{1.9}
& $0.9\times$
& \pmc{17.4}{0.7}
& \pmc{64.8}{0.9}
& \pmc{58.6}{0.7}
& \pmc{21.4}{1.4}
& \pmc{\underline{40.6}}{0.5}
& \pmc{\textbf{7.4}}{2.5}
& $\mathbf{1.8\times}$ \\

\textbf{\textsc{METIS}}
& \pmc{20.4}{1.0}
& \pmc{9.3}{0.7}
& \pmc{64.8}{0.4}
& \pmc{20.6}{1.3}
& \pmc{\textbf{28.8}}{0.5}
& \pmc{\underline{10.6}}{3.8}
& $\underline{1.1\times}$
& \pmc{18.3}{0.6}
& \pmc{66.1}{0.7}
& \pmc{59.5}{0.5}
& \pmc{21.3}{1.3}
& \pmc{\textbf{41.3}}{0.5}
& \pmc{\underline{8.3}}{2.7}
& $\underline{1.6\times}$ \\

\midrule

% =========================================================
% Block 3
% =========================================================
\multicolumn{1}{@{}l}{\textbf{Model}}
& \multicolumn{7}{c}{\textbf{Llama-3.1-8B-Instruct}}
& \multicolumn{7}{c}{\textbf{DS-R1-Distill-Llama-8B}} \\

\cmidrule(lr){2-8}
\cmidrule(lr){9-15}

No Curr.
& \pmc{14.8}{1.2}
& \pmc{6.4}{0.8}
& \pmc{50.1}{2.1}
& \pmc{18.9}{2.8}
& \pmc{22.6}{0.9}
& \pmc{8.4}{3.3}
& $1.0\times$
& \pmc{39.2}{0.5}
& \pmc{79.9}{0.6}
& \pmc{64.3}{0.3}
& \pmc{13.6}{1.1}
& \pmc{\underline{49.3}}{0.2}
& \pmc{18.2}{6.6}
& $1.0\times$ \\

SEC
& \pmc{16.6}{0.8}
& \pmc{7.6}{0.8}
& \pmc{49.9}{2.8}
& \pmc{16.2}{1.0}
& \pmc{22.6}{0.8}
& \pmc{\textbf{5.5}}{3.1}
& $\mathbf{1.5\times}$
& \pmc{38.0}{0.3}
& \pmc{77.4}{0.6}
& \pmc{65.7}{0.2}
& \pmc{12.6}{1.2}
& \pmc{48.4}{0.4}
& \pmc{17.2}{4.5}
& $1.1\times$ \\

ADCL
& \pmc{16.9}{1.0}
& \pmc{6.8}{0.8}
& \pmc{50.3}{0.7}
& \pmc{19.0}{0.9}
& \pmc{23.2}{0.5}
& \pmc{24.3}{4.2}
& $0.3\times$
& \pmc{36.9}{0.2}
& \pmc{77.6}{0.9}
& \pmc{66.4}{1.3}
& \pmc{12.0}{1.9}
& \pmc{48.2}{0.9}
& \pmc{34.0}{8.3}
& $0.5\times$ \\

PCL
& \pmc{17.0}{1.0}
& \pmc{8.1}{0.8}
& \pmc{49.5}{1.2}
& \pmc{22.3}{1.2}
& \pmc{\textbf{24.2}}{0.6}
& \pmc{19.0}{4.0}
& $0.4\times$
& \pmc{38.6}{0.4}
& \pmc{77.4}{1.2}
& \pmc{68.1}{0.8}
& \pmc{12.8}{1.6}
& \pmc{49.2}{0.7}
& \pmc{\underline{14.4}}{3.8}
& $\underline{1.3\times}$ \\

\textbf{\textsc{METIS}}
& \pmc{15.2}{1.2}
& \pmc{9.1}{0.7}
& \pmc{49.8}{0.7}
& \pmc{19.4}{0.9}
& \pmc{\underline{23.4}}{0.4}
& \pmc{\underline{6.8}}{2.9}
& $\underline{1.2\times}$
& \pmc{39.8}{1.1}
& \pmc{78.9}{0.7}
& \pmc{65.9}{0.5}
& \pmc{14.9}{0.8}
& \pmc{\textbf{49.9}}{0.2}
& \pmc{\textbf{10.5}}{3.1}
& $\mathbf{1.7\times}$ \\

\bottomrule
\end{tabular}%
}
% \vspace{-3.5mm}

\end{table*}

% \textbf{Evaluation protocol.} For main results, we conduct three full training runs with different random seeds. For each run, we select the checkpoint with the highest aggregated average pass@1 score over all benchmarks as the best checkpoint. Tab.~\ref{tab:main_results} reports the mean and std of the pass@1 achieved by the best checkpoint, and the wall-clock time to reach it. Training dynamics are shown in Fig.~\ref{fig:main_curves}. 

% Given the small size of AIME24/25, their pass@1 estimates are sensitive to sampling variance; we therefore report avg@10 for these two benchmarks.

\subsection{Main Results on METIS Performance}
\label{mainexp}
\begin{figure}[t]
    \centering
\includegraphics[width=0.80\linewidth]{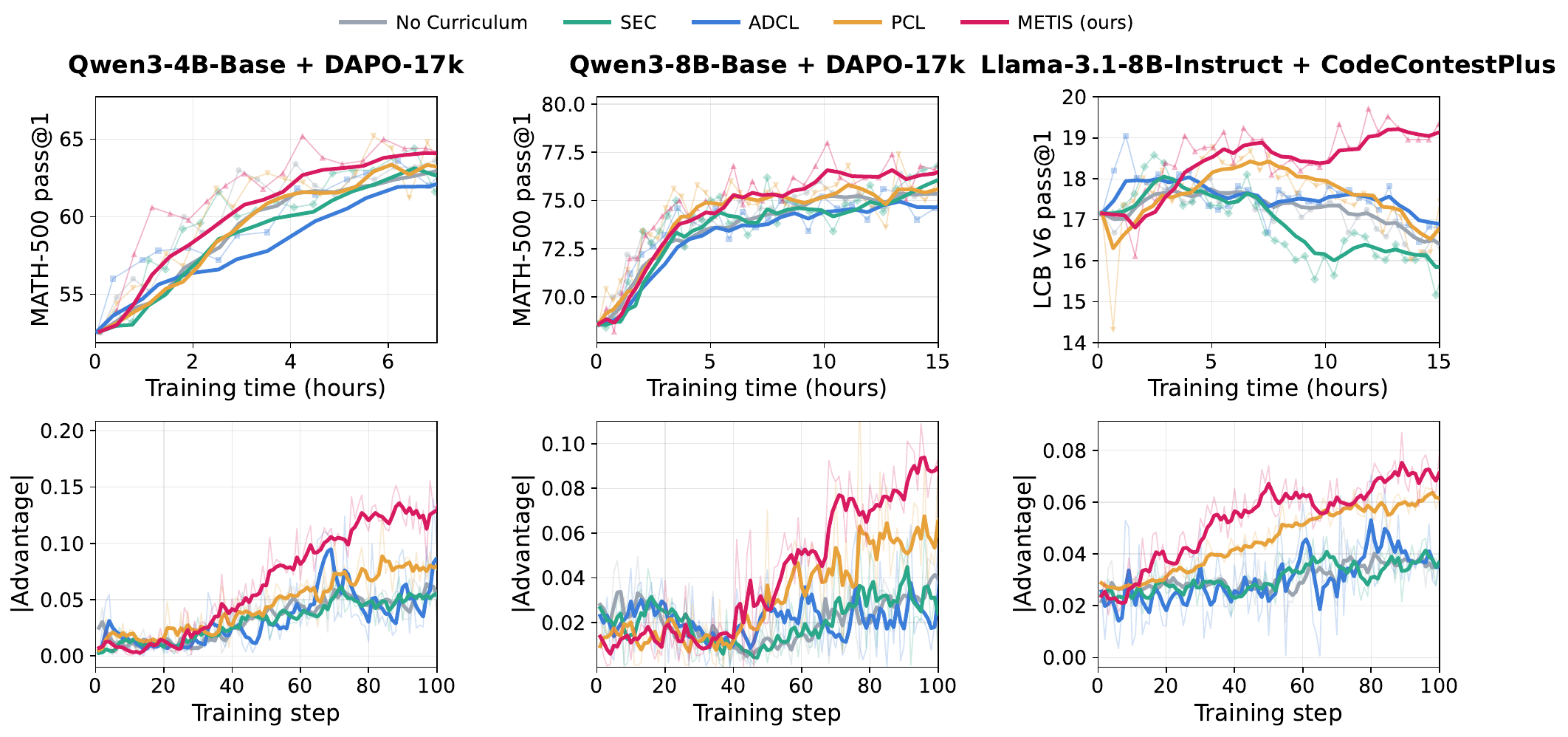}
    % \vspace{-2mm}
    \caption{Training dynamics of \textsc{METIS}. \textbf{Top:} downstream pass@1 versus wall-clock training time. \textbf{Bottom:} mean magnitude of the group-relative advantage \(|A|\) per training step. 
    % \textsc{METIS} reaches higher pass@1 earlier while sustaining a larger per-step learning signal than all baselines.
    }
    \label{fig:main_curves}
    % \vspace{-5pt}
\end{figure}

We run each experiment with three random seeds and select, for each run, the checkpoint with the highest average Pass@1 across all benchmarks. Tab.~\ref{tab:main_results} reports the mean and standard deviation of the selected checkpoints' Pass@1 scores and the wall-clock time required to reach them. Fig.~\ref{fig:main_curves} shows the training dynamics. Results on agentic benchmarks are provided in Tab.~\ref{tab:bfcl_v3} (in App.~\ref{app:bfcl_extension}).

\textbf{(i) Higher accuracy with less training time.}
\textsc{METIS} achieves the highest average accuracy in nearly every row of Tab.~\ref{tab:main_results} and reaches its best checkpoint up to \(2.1\times\) faster than No Curriculum. It outperforms the strongest baseline, PCL, without an auxiliary scorer. SEC trains the fastest in several settings but consistently trails on accuracy since its category-level re-weighting misses the prompt-level informativeness signal. Finally, ADCL is the costliest baseline, since its offline difficulty re-estimation pass adds substantial overhead and its fixed schedule does not co-evolve with the policy.

\textbf{(ii) Faster convergence.}
Fig.~\ref{fig:main_curves} (top row) plots pass@1 versus training time. \textsc{METIS} reaches the same level of accuracy earlier in training and continues improving thereafter, indicating that gains come from steady accumulation of learning signal rather than a one-time selection effect.

\textbf{(iii) Larger learning signal per step.}
Fig.~\ref{fig:main_curves} (bottom row) tracks the mean magnitude of the group-relative advantage \(|A|\), which is the proxy for the prompt informativeness \(v_\theta(x)\) defined in Eq.~\eqref{eq:reward_variance}. \textsc{METIS} sustains substantially higher \(|A|\) throughout training, indicating that its curriculum selection consistently allocates updates to more informative prompts.

\textbf{Computational costs.} Fig.~\ref{fig:overhead} reports the per-step wall-clock increase (left) and relative throughput (tokens/s/GPU) drop (right), relative to No Curriculum. \textsc{METIS} adds only a short judgment pass over \(m\cdot B\) candidates, yielding marginal costs on both metrics. In contrast, ADCL incurs wall-clock overhead from extra difficulty-estimation rollouts, while PCL reduces throughput by training a separate value model. SEC is cheapest but sacrifices accuracy through coarse category-level curriculum. \textsc{METIS} is a unique prompt-level method that remains efficient on both axes, providing effective curriculum at near-negligible cost.
% Thus, \textsc{METIS} uniquely combines prompt-level selection with low overhead on both metrics.

\subsection{Ablation and Hyper-parameter Studies}
\label{sec:mainablation}

\textbf{Ablations on curriculum judgment internalization.}
To validate our claim on internalized judgment, in Tab.~\ref{tab:internal_external}, we ablate the judgment provider and judgment learning on Qwen3-4B-Base with DAPO-17k. For \textsc{METIS}, the internal variant uses the task policy as the ICL judge, whereas the external variant uses a separate, identically initialized model; replacing ICL with an auxiliary value head yields the PCL-style variants. Internalization raises \textsc{METIS}'s average score from 27.3 to 28.8, while judgment learning further improves performance in both settings. Moreover, even without judgment supervision, internal \textsc{METIS} outperforms its supervised external variant (28.4 vs.\ 27.3), suggesting that task learning transfers to judgment through the shared backbone. Notably, internalizing PCL's value head lowers performance from 28.2 to 26.6, suggesting that token-based ICL may be more compatible with policy optimization than auxiliary scalar prediction using an MSE loss.

\begin{table}[t]
\centering
\caption{Effects of curriculum judgment internalization on Qwen3-4B-Base + DAPO-17k. \textbf{Internal}: the task-solving policy produces the curriculum judgment; \textbf{External}: a separate duplicated model produces the curriculum judgment. \textbf{Judgment Learning}: whether the judge is updated from realized variances (\(\mathcal{L}_{\mathrm{judge}}\) for \textsc{METIS}; MSE regression for PCL's value head). }
\label{tab:internal_external}
% \vspace{2pt} 
\scriptsize
\setlength{\tabcolsep}{4pt}
\renewcommand{\arraystretch}{1.05}
\resizebox{0.8\textwidth}{!}{%
\begin{tabular}{l l c cccc c}
\toprule
Method & Judgement Provider & Judgment Learning & AIME24 & AIME25 & MATH500 & Minerva & Avg.\,$\uparrow$ \\
\midrule
\textsc{METIS} (default)
& Internal (ICL) & \cmark
& 20.4 $\pm$ 1.0
& 9.3 $\pm$ 0.7
& 64.8 $\pm$ 0.4
& 20.6 $\pm$ 1.3
& \textbf{28.8} $\pm$ 0.5 \\

\textsc{METIS}
& Internal (ICL) & \xmark
& 20.1 $\pm$ 1.1
& 10.1 $\pm$ 0.8
& 63.3 $\pm$ 0.8
& 20.2 $\pm$ 1.0
& 28.4 $\pm$ 0.6 \\

\textsc{METIS}
& External (ICL) & \cmark
& 19.0 $\pm$ 0.9
& 7.2 $\pm$ 0.8
& 63.6 $\pm$ 1.0
& 19.4 $\pm$ 0.9
& 27.3 $\pm$ 0.6 \\

\textsc{METIS}
& External (ICL) & \xmark
& 18.8 $\pm$ 0.8
& 6.4 $\pm$ 0.7
& 64.0 $\pm$ 0.8
& 18.5 $\pm$ 0.8
& 26.9 $\pm$ 0.5 \\

\midrule

PCL (default)
& External (value head) & \cmark
& 18.4 $\pm$ 1.5
& 11.4 $\pm$ 1.1
& 65.1 $\pm$ 0.1
& 17.8 $\pm$ 0.8
& 28.2 $\pm$ 0.5 \\

PCL
& Internal (value head) & \cmark
& 15.6 $\pm$ 1.0
& 8.2 $\pm$ 0.8
& 64.7 $\pm$ 1.0
& 18.0 $\pm$ 0.7
& 26.6 $\pm$ 0.6 \\
\bottomrule
\end{tabular}%
}
\end{table}

\begin{figure}[t]
\centering
\begin{minipage}[t]{0.48\textwidth}
  \centering
  \tiny
  \setlength{\tabcolsep}{1pt}
  \renewcommand{\arraystretch}{0.9}
  \captionof{table}{Effects of METIS designs on Qwen3-4B-Base + DAPO-17k, comparing full \textsc{METIS} (\(K{=}3,\,\lambda{=}0.01\)) against removing in-context evidence (\(K{=}0\)) and the joint judgment loss (\(\lambda{=}0\)), and baselines (No Curriculum and PCL).}
  \label{tab:ablation_components}
  \vspace{2pt}
  \resizebox{\linewidth}{!}{%
  \begin{tabular}{l cccc | c | c}
  \toprule
  Method & AIME24 & AIME25 & MATH500 & Minerva & Avg.\,$\uparrow$ & Fail.\,(\%)\,$\downarrow$ \\
  \midrule
   No Curriculum                          & 17.0 & 10.3  & 66.4 & 16.2 & 27.5 & ---  \\
  PCL                                    & 19.3 & 10.7  & 65.0 & 18.4 & 28.3 & ---  \\
  \midrule
  w/o ICL                                & 17.0 & 7.0  & 64.6 & 21.0 & 27.4 & 87.6 \\
  w/o \(\mathcal{L}_{\mathrm{judge}}\)   & 18.0 & 9.7 & 64.4 & 20.2 & 28.1 & 2.2  \\
    \textsc{METIS}                         & 19.3 & 9.3 & 65.2 & 21.7 & \textbf{28.9} & \textbf{0.3}  \\
  \bottomrule
  \end{tabular}%
  }
\end{minipage}%
\hfill
\begin{minipage}[t]{0.45\textwidth}
  \centering
  \vspace{0pt}% align top with neighbor
  % \vspace{-2pt}
\includegraphics[width=\linewidth]{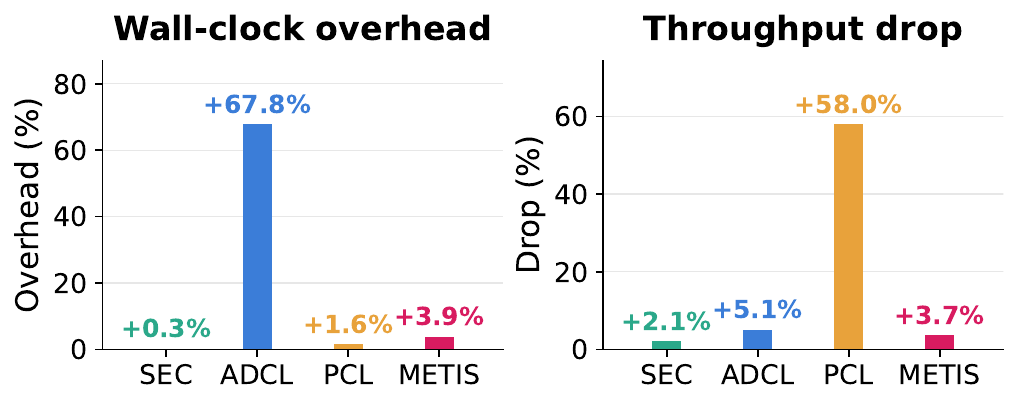}
% \vspace{1pt}
  \captionof{figure}{Compute overhead of curriculum methods, measured against No Curriculum. \textbf{Left:} wall-clock overhead. \textbf{Right:} per-step throughput drop. Lower is better on both axes.}
  \label{fig:overhead}
\end{minipage}
\end{figure}

\textbf{Ablations on METIS component designs.} Tab.~\ref{tab:ablation_components} compares full \textsc{METIS} against two reduced variants, reporting downstream performance and the \emph{failure rate}, defined as the percentage of self-judgment responses from which a valid $\hat v_\theta(x)$ cannot be parsed. When w/o ICL, the parse-failure rate for \(\hat{v}_\theta(x)\) rises from \(0.3\%\) to \(87.6\%\), while average Pass@1 declines. Without in-context examples, the policy attempts to solve prompts rather than estimate their informativeness (App.~\ref{app:no_icl_example}). When w/o  \(\mathcal{L}_{\mathrm{judge}}\), METIS becomes open-loop, and the performance also drops. As  Fig.~\ref{fig:judgment_diagnostics} (in App.~\ref{app:component_ablations}) shows, the loss steadily improves \(R_{\mathrm{judge}}\) (left), aligns the mean \(\hat{v}_\theta(x)\) with the competence frontier (middle), and preserves differentiated predictions (right); without it, the latter two collapse into a flat mean and clustered predictions. Even so, this open-loop variant still beats the baselines in Tab.~\ref{tab:ablation_components}, showing that ICL alone is effective while \(\mathcal{L}_{\mathrm{judge}}\) closes the loop and adds a further accuracy gain. 

\textbf{Hyperparameter choices.}
We study four \textsc{METIS} hyperparameters (Tab.~\ref{tab:ablation_lambda_k} in App.~\ref{app:hyperparameter}). Performance is similar for \(\lambda\in\{0.01,0.1\}\), whereas larger values hurt accuracy; at \(\lambda=1\), predictions collapse toward \(0.25\) (Fig.~\ref{fig:pool_predictions}). Setting \(K=0\) causes frequent judgment failures, \(K=3\) nearly eliminates them, and \(K=9\) provides no further gain. Performance is stable across pool multipliers \(m\in\{4,8,12\}\). Finally, the EMA rate \(\alpha\in\{0.5,0.95\}\) performs similarly, while \(\alpha=0\) performs worse, indicating that EMA smoothing stabilizes judgment updates.

\subsection{Behavior Analyses and Discussions}
\label{sec:discussion}

\begin{figure}[t]
    \centering
\includegraphics[width=1\linewidth]{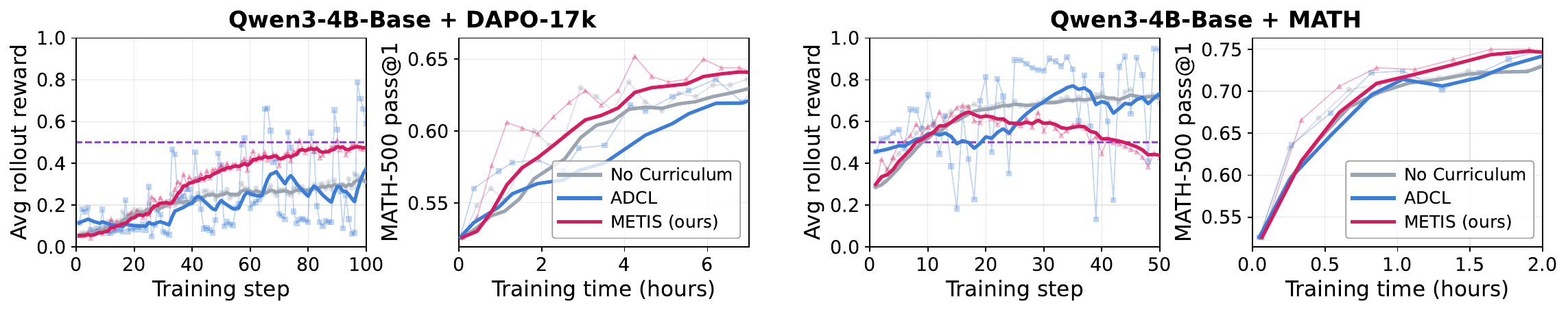}
% \vspace{-3mm}
    \caption{The average rollout reward and the corresponding training performance (pass@1) curve on a hard task (DAPO) and an easy task (MATH). Purple dashed line marks the maximum-informativeness \(p=0.5\). \textsc{METIS} regulates around \(p=0.5\) from both sides, leading to higher accuracy.}
    \label{fig:dynamic_showcase}
\end{figure}

\textbf{\textsc{METIS} tracks the competence frontier.}
To test whether \textsc{METIS} adapts its curriculum to the policy's evolving competence, Fig.~\ref{fig:dynamic_showcase} tracks the mean rollout reward of the prompts selected during training. For binary rewards, this mean equals the empirical pass rate \(p\), and the reward variance \(p(1-p)\) is maximized at \(p=0.5\). Thus, an adaptive curriculum should move selected prompts toward this informative frontier as the policy evolves. We examine two contrasting regimes: DAPO-17k, whose prompts are initially challenging, and MATH, which the policy masters more quickly. On DAPO-17k, all methods start near \(p=0\), but \textsc{METIS} approaches \(0.5\) fastest, shifting from consistently failed to intermittently solved prompts. On MATH, \(p\) initially rises above \(0.5\); \textsc{METIS} then shifts toward harder prompts and returns near the frontier, while the baselines generally drift upward. These trajectories suggest that \textsc{METIS} tracks the informative region from both directions, coinciding with higher pass@1. In contrast, No Curriculum remains closer to the extremes, while ADCL fluctuates around \(p=0.5\) without consistently following the moving frontier.

\begin{wraptable}{r}{0.42\textwidth}
\vspace{-12pt}
\centering
\caption{Unique prompt coverage rate.}
\label{tab:maincoverage}
\footnotesize
\setlength{\tabcolsep}{5pt}
\renewcommand{\arraystretch}{1.05}
\resizebox{0.65\linewidth}{!}{%
\begin{tabular}{@{}lrr@{}}
\toprule
Setting & PCL & \textsc{METIS} \\
\midrule
Qwen3-4B/DAPO & 39\% & \textbf{88\%} \\
Qwen3-8B/DAPO & 46\% & \textbf{82\%} \\
Llama-8B/Code & 43\% & \textbf{91\%} \\
\bottomrule
\end{tabular}%
}
\vspace{-0.4\baselineskip}
\end{wraptable}
\textbf{METIS ensures curriculum coverage.}
To test whether self-selection collapses to a fixed subset, we measure unique prompt coverage during training, in Tab.~\ref{tab:maincoverage}. Across settings, \textsc{METIS} covers \(82\)-\(91\%\) of the dataset, versus \(39\)-\(46\%\) for PCL. This suggests that \textsc{METIS}'s adaptive frontier tracking ensures coverage, while PCL's external judge may drift toward a narrower subset.

\textbf{\textsc{METIS} increases training stability.}
We quantify local training stability using the detrended standard deviation of the pass@1 trajectory, which captures step-to-step fluctuations after removing the overall learning trend (Tab.~\ref{tab:stability} in App.~\ref{app:coveragestablity}). \textsc{METIS} exhibits smoother training dynamics than PCL (e.g., \(0.92\) vs.\ \(1.14\)). Its higher late-stage policy entropy (Tab.~\ref{tab:entropy} in App.~\ref{app:coveragestablity}) further suggests that \textsc{METIS} is less prone to entropy collapse during training.

%% ============================================================================
\section{Conclusions}
\label{sec:conclusion}
%% ============================================================================

We presented \textsc{METIS}, a closed-loop framework in which the policy itself predicts prompt informativeness via in-context self-judgment and is reinforced from realized rollout outcomes, learning both to solve prompts and to decide which prompts to learn from next. Across math, code, and agentic tasks, \textsc{METIS} delivers higher accuracy with up to 2.1$\times$ training speedups over wall-clock time at only \(3.9\%\) per-step overhead. These results indicate that prompt-level curriculum judgment can be reliably absorbed into the policy itself, removing the need for hand-tuned schedules or separately trained selectors in RFT. One potential limitation is that \textsc{METIS} relies on the base policy's in-context learning capacity, which weaker backbones may not meet. We provide a detailed discussion on such failure boundaries in App.~\ref{app:failure_boundary}. Future work includes studying whether learned curriculum judgment transfers across tasks and domains, and deploying \textsc{METIS} on real-world LLM RFT systems. Our code, together with reproducibility instructions, will be made publicly available upon acceptance.

\newpage
\section*{AI use statement}
In this work, we used generative AI tools only to polish the writing and improve the clarity and readability of the manuscript. We did not use generative AI tools for other activities requiring disclosure under the ICLR 2027 AI Policy for Authors. The formulation of the research problem, development of the method, experimental design, analysis of results, and interpretation of findings are entirely the authors' own work. We take full responsibility for the content of this paper.
% \newpage
\section*{Ethics Statement}

This work does not involve human subjects, personally identifiable information, or sensitive personal data. 
The experiments are conducted on publicly available benchmarks and pretrained models. 
We do not identify any immediate ethical concerns specific to the proposed methodology beyond those generally 
associated with the development and deployment of large language models. 
The authors take responsibility for ensuring that the research and experimental procedures comply with 
applicable institutional and licensing requirements.

\section*{Reproducibility Statement}
We provide detailed descriptions of the proposed method, training procedure, hyperparameters, prompts, 
evaluation protocol, and ablation settings in the main paper and appendix.  Upon acceptance, we will publicly release the source code to facilitate reproducibility and future research.

{
\small
\bibliographystyle{iclr2027_conference}
\bibliography{references}
}

%%%%%%%%%%%%%%%%%%%%%%%%%%%%%%%%%%%%%%%%%%%%%%%%%%%%%%%%%%%%
\newpage

\appendix

% \setcounter{page}{1} %!!!!!!!!!!!!!!!!!!!!

% \vbox{
% \hsize\textwidth
% \hrule height 4pt
% \vskip 0.25in%
% \vskip -\parskip%
% \centering
% {\LARGE\bf Internalizing Curriculum Judgment for
% LLM Reinforcement Fine-Tuning (Appendix) \par}
% \vskip 0.29in
% \vskip -\parskip
% \hrule height 1pt
% \vskip 0.3in%
% %   \textbf{Paper ID: 9954}
% % \vskip 0.3in
% }

% Re-enable TOC entries (sections + subsections) for the appendix and print a
% hyperlinked table of contents on the first appendix page.
\addtocontents{toc}{\protect\setcounter{tocdepth}{2}}
\renewcommand{\contentsname}{Appendix Contents}
\tableofcontents
\newpage

\section{Additional Experimental Details}
\label{app:details}
%% ============================================================================

\subsection{Prompt template for the self-judgment}
\label{app:prompt_template}

Box~\ref{box:variance_prompt} shows the full prompt template the policy reads when producing its self-judgment for a candidate prompt. The template comprises a fixed system message defining the variance-prediction task and a per-candidate user message containing the in-context exemplars from $\mathcal{M}_t$ followed by the candidate prompt.

\begin{promptbox}[Self-judgment prompt for variance prediction ]
\label{box:variance_prompt}
\ttfamily\scriptsize

\textbf{System message.}\\[3pt]
You are a predictor that estimates the reward variance for a candidate problem. Reward variance is the variance of the rewards across multiple solution attempts by the current model on the same problem. High variance means rollouts disagree --- some attempts score well and others poorly --- which is exactly the regime where GRPO has the strongest learning signal. Low variance means rollouts agree (the model consistently passes or consistently fails), so there is little gradient to extract.\\[3pt]
The labeled examples show actual reward variances from recent model performance; use them to calibrate your prediction.\\[3pt]
Output your final prediction inside \verb|\boxed{}|, choosing one number from this list: \verb|[0.00, 0.02, 0.04, 0.06, 0.08, 0.10, 0.12, 0.15, 0.18, 0.20, 0.25]|. Example final line: \verb|\boxed{0.10}|.\\[8pt]

\textbf{User message.}\\[3pt]
Predict the reward variance for the next problem. The examples below show actual reward variances from recent model performance. Use them to calibrate your prediction. Put your final variance prediction inside \verb|\boxed{}|.\\[6pt]

\verb|[Example 1]|\\
\verb|PROMPT:|\\
\verb|"""|\\
\textit{$\langle$exemplar prompt 1$\rangle$}\\
\verb|"""|\\
\verb|REWARD_VARIANCE: 0.000|\\[4pt]
\hspace*{1em}$\vdots$\\[4pt]
\verb|[Example K]|\\
\verb|PROMPT:|\\
\verb|"""|\\
\textit{$\langle$exemplar prompt K$\rangle$}\\
\verb|"""|\\
\verb|REWARD_VARIANCE: 0.250|\\[6pt]

\verb|[Candidate]|\\
\verb|PROMPT:|\\
\verb|"""|\\
\textit{$\langle$candidate prompt$\rangle$}\\
\verb|"""|

\end{promptbox}

\paragraph{Calibration memory}
\label{app:memory_details}
After each training step, every prompt selected at that step is written to the training history together with its realized rollout reward variance, replacing the previous step's contents. The history therefore always holds the full selected batch of the most recent step, and the stored pairs reflect the current policy's behavior. At the next step, the \(K\) in-context exemplars are retrieved from the training history by a deterministic nearest-target rule: for each reference value in \(\{0.00,\,0.10,\,0.25\}\), representing low, moderate, and high variance, we pick the \(K/3\) prompts from training history whose realized variance is closest to that reference, without replacement. The exemplars are always presented in the same low-to-high order across all experiments. We expect order sensitivity to be limited here because each exemplar carries an explicit numeric label (\texttt{REWARD\_VARIANCE}), so the mapping is given rather than inferred from position.

\subsection{Dataset Pre-processing}
\label{app:dataset_preprocessing}

\paragraph{DAPO-Math-17k (math, training).}
We use the released DAPO-17k collection of $17.4$k competition-style math problems~\citep{yu2026dapo} as is, without additional filtering. Each problem ships with a short reference final answer. The policy is instructed to produce a full chain-of-thought solution and a final-answer line; the reward is binary, equal to one when the extracted final answer exactly matches the reference under standard math normalization, and zero otherwise.

\paragraph{CodeContests-Plus (code, training).}
We start from the public CodeContests-Plus dataset~\citep{wang2025codecontestshighqualitytestcase} and apply three filters that together leave $4{,}432$ problems. \emph{First}, we drop any problem with no Python program among its reference correct submissions, since we train a Python-only policy and the per-problem time limits are tuned for C++. \emph{Second}, we drop problems with more than $25$ test cases. \emph{Third}, we drop problems whose statements reference figures (images or vector diagrams) that a text-only policy cannot read. The training reward is the fraction of test cases passed by the model's program when executed, continuous in $[0,1]$. The statement seen by the self-judgment retains only the prose part of the description: we strip the formal Constraints, Input, Output, Examples, and Notes, as these are uninformative for variance prediction and would otherwise crowd the calibration context.

\paragraph{BFCL v3 (agentic, training).}
We use all five multi-turn categories of the Berkeley Function Calling Leaderboard v3~\citep{patil2025bfcl} -- Base, Missed-Function, Missed-Parameter, Long-Context, and Composite. Each task is a sequence of user turns interleaved with API calls against a stateful environment.  The reference tool-call trajectory serves as ground truth, and the reward is computed against the actual sequence of tool calls executed by the model, yielding a continuous score in $[0,1]$. We split the resulting tasks $80\%/20\%$ into train and validation with per-category stratification, ensuring every category appears in both splits.

\paragraph{Math evaluation suite.}
AIME~2024, AIME~2025, MATH-500, and Minerva Math share a single preparation pipeline. To keep offline numbers directly comparable to the training-time validation curves, we reuse the exact problem-wrapping instruction (think step by step and place the final answer in a boxed expression) together with the exact answer-extraction and Hendrycks-MATH normalization used during training. AIME~2024 contributes $30$ problems and AIME~2025 contributes $30$ problems combining the two AIME~2025 olympiad sittings; MATH-500 contributes $500$ problems whose released answers are already short strings. Minerva Math contributes $272$ problems whose release contains full derivations rather than short answers. 

\paragraph{Code evaluation suite.}
The code evaluation suite combines LiveCodeBench v6~\citep{jain2025livecodebench} ($1{,}055$ problems), HumanEval ($164$ problems), MBPP (full test split, $974$ problems), and BigCodeBench (v0.1.4, $1{,}140$ problems). LiveCodeBench v6 is a competitive-programming benchmark with stdin/stdout test cases; we evaluate it under the same stdin/stdout prompt and execution-based scorer used during CodeContests-Plus training. The remaining three benchmarks are function-call-based. Scoring on all four benchmarks is performed by executing the extracted code against the benchmark's official tests in a sandbox, with the score equal to the fraction of tests passed.

\subsection{Training Infrastructure}
\label{app:training_infra}

All experiments are implemented on the VERL reinforcement-learning framework~\citep{sheng2025hybridflow} with vLLM~\citep{kwon2023efficientmemorymanagementlarge} as the policy rollout engine. For mathematical-reasoning training, rewards are cheap to compute (a verifier match against a short reference answer), so rollout, reward, and policy update run on a fully synchronous schedule. For code-generation and agentic function-calling training, rewards are dominated by external execution; we therefore run reward computation asynchronously, overlapping the scoring of completed rollouts with continued generation so that GPUs stay busy on the rollout engine while CPU workers handle the heavier execution side.

\subsection{Training hyperparameters}
\label{app:training_hparams}

Tab.~\ref{tab:grpo_hparams} reports the GRPO training hyperparameters used in our main experiments. Within each task family, all methods (No Curriculum, SEC, ADCL, PCL, and \textsc{METIS}) share these values; method-specific knobs are reported in Sec.~\ref{app:method_hparams}. The KL-regularization term and the entropy bonus are both disabled across all task families. Following~\citet{liu2025understandingr1zeroliketrainingcritical}, we drop the per-prompt standard-deviation normalization from the GRPO advantage. Each training run is capped by wall-clock time rather than by a fixed number of epochs: $40$ hours for $8$B backbones and $20$ hours for $4$B backbones. Within that budget we always report the best validation checkpoint reached during training along with the wall-clock time at which it is attained. All runs execute on a cluster of $12$ nodes, each with $8\times$NVIDIA H200 GPUs. The full set of experiments reported in this paper consumed approximately $80$K H200 GPU-hours. 

\begin{table}[h]
\centering\small
\setlength{\tabcolsep}{12pt}
\renewcommand{\arraystretch}{1.15}
\caption{Shared GRPO training hyperparameters.}
\label{tab:grpo_hparams}
\begin{tabular}{lccc}
\toprule
Hyperparameter             & Math (DAPO-17k)    & Code (CCP)         & Agentic (BFCL v3)  \\
\midrule
Train batch size           & $256$              & $128$              & $64$               \\
PPO mini-batch size        & $64$               & $64$               & $64$               \\
Micro-batch per GPU        & $16$               & $8$                & $32$               \\
Learning rate              & $2{\times}10^{-6}$ & $2{\times}10^{-6}$ & $1{\times}10^{-6}$ \\
PPO clip ratio             & $0.2$              & $0.2$              & $0.2$              \\
\addlinespace[2pt]
Rollouts per prompt        & $8$                & $8$                & $8$                \\
Max prompt length          & $2048$             & $2048$             & $2048$             \\
Max response length        & $8192$             & $8192$             & $4096$             \\
\midrule
Optimizer                  & \multicolumn{3}{c}{AdamW (no learning-rate warmup or decay)} \\
Mixed precision            & \multicolumn{3}{c}{bfloat16 parameters and gradient reductions; fp32 optimizer buffers} \\
Gradient checkpointing     & \multicolumn{3}{c}{enabled} \\
Model sharding             & \multicolumn{3}{c}{FSDP} \\
\bottomrule
\end{tabular}
\end{table}

\subsection{Baseline implementation details}
\label{app:baselines}

We compare \textsc{METIS} against three external-curriculum baselines: SEC, ADCL, and PCL. All three were originally proposed for binary-reward mathematical reasoning. We re-implement each on top of our shared GRPO trainer so that every method sees identical rollouts, optimizer, and hardware; the only differences are the curriculum mechanism itself and, for ADCL and PCL, the modifications described below that extend them to continuous-reward tasks.

\paragraph{SEC.}
Self-Evolving Curriculum~\citep{chen2025self} treats curriculum learning as a non-stationary multi-armed bandit over coarse prompt categories. Our implementation closely follows the authors' released code. We partition the training set into ten categories. At each training step, the sampler draws a category via Boltzmann softmax over per-category Q-values and then samples prompts uniformly within the chosen category until the batch is full. After the GRPO update, the Q-value of every category appearing in the batch is updated by an exponential moving average toward the mean per-token absolute advantage observed on its prompts, so categories that currently carry the largest gradient signal are progressively up-weighted. Because SEC consumes the GRPO advantage signal directly and never inspects the raw reward, the algorithm transfers to continuous-reward tasks (code generation, agentic function calling) without modification.

\paragraph{ADCL.}
Adaptive Difficulty Curriculum Learning~\citep{zhang2025learning} is an offline-then-online easy-to-hard curriculum. Our implementation strictly follows the algorithm description in the original paper, using the curriculum-only variant. Before training, the base policy is rolled out on the entire training set to assign each prompt an initial difficulty score; the dataset is then sorted in ascending order by difficulty and partitioned into a fixed number of sequential chunks (eight in our runs). Training proceeds chunk by chunk; between chunks, the difficulty of the next chunk is re-estimated under the updated policy and the chunk is locally re-sorted, so the curriculum tracks the model's evolving competence without re-sorting the whole dataset.

In the binary-reward setting (mathematical reasoning), we keep the original definition of difficulty as one minus the empirical pass rate. For continuous-reward tasks (code generation, BFCL), pass rate is undefined since rewards are no longer $0/1$; to extend ADCL to this regime, we rank prompts by their empirical \emph{reward variance} instead: low-variance prompts are treated as easy, and high-variance prompts as hard. The chunk structure and between-chunk re-estimation are otherwise unchanged.

\paragraph{PCL.}
Prompt Curriculum Learning~\citep{gao2026prompt} replaces a hand-coded curriculum with a learned auxiliary value model. Our implementation strictly follows the algorithm description in the original paper. At each training step we sample a candidate pool several times larger than the batch, score every candidate with the value model, select the prompts whose predicted score best matches the desired target, run GRPO on the selected batch, and update the value model from the realized outcomes of that step. The value model is itself a language model: we initialize its backbone from the same pretrained model as the policy and attach a small scalar value head on top of the prompt's final hidden state. It thus inherits the policy's representation of natural-language problems while learning a separate readout for the curriculum signal. The value model and the policy do not share parameters during training.

In the binary-reward setting we follow the original PCL: the value head regresses (under MSE) onto the empirical pass rate of each prompt, and selection picks the batch whose predicted pass rate is closest to a target of $0.5$. For continuous-reward tasks, where pass rate is no longer defined, we adapt PCL to the informativeness signal used throughout this paper: the value head instead regresses onto the empirical \emph{reward variance} of each prompt, and selection picks the prompts with the \emph{highest} predicted variance. This keeps the auxiliary-selector design of PCL intact while making it apply uniformly to both binary and continuous rewards.

\subsection{Method-specific hyperparameters}
\label{app:method_hparams}

Tab.~\ref{tab:method_hparams} lists the hyperparameters specific to each curriculum method, i.e., the knobs not shared with the GRPO defaults of Tab.~\ref{tab:grpo_hparams}. Within each method, the same values are used across all backbones and tasks unless otherwise noted.

\begin{table}[h]
\centering
\small
\caption{Method-specific hyperparameters for \textsc{METIS} and the three baselines.}
\label{tab:method_hparams}

\begin{minipage}[t]{0.48\linewidth}
\centering
\textbf{\textsc{METIS}}\\[3pt]
\begin{tabular}{lc}
\toprule
Hyperparameter & Value \\
\midrule
Judgment loss weight             & $0.01$ \\
Number of in-context examples    & $3$    \\
Baseline EMA rate                & $0.95$ \\
Pool multiplier                  & $8$    \\
\bottomrule
\end{tabular}
\end{minipage}%
\hfill
\begin{minipage}[t]{0.48\linewidth}
\centering
\textbf{SEC}\\[3pt]
\begin{tabular}{lc}
\toprule
Hyperparameter & Value \\
\midrule
Number of categories          & $10$  \\
EMA rate                      & $0.5$ \\
Boltzmann temperature         & $0.4$ \\
Pool multiplier               & $8$   \\
\bottomrule
\end{tabular}
\end{minipage}

% \vspace{0.9em}

\begin{minipage}[t]{0.48\linewidth}
\centering
\textbf{ADCL}\\[3pt]
\begin{tabular}{lc}
\toprule
Hyperparameter & Value \\
\midrule
Number of chunks                          & $8$ \\
Rollouts per prompt for difficulty        & $4$ \\
\bottomrule
\end{tabular}
\end{minipage}%
\hfill
\begin{minipage}[t]{0.48\linewidth}
\centering
\textbf{PCL}\\[3pt]
\begin{tabular}{lc}
\toprule
Hyperparameter & Value \\
\midrule
Pool multiplier              & $8$                \\
Value-head hidden dimension  & $512$              \\
Value-model learning rate    & $1{\times}10^{-4}$ \\
Value-model batch size       & $64$               \\
Updates per training step    & $1$                \\
Selection target (binary)    & $0.5$              \\
\bottomrule
\end{tabular}
\end{minipage}

\end{table}

% \newpage

%% ============================================================================
\section{Extended Results}
\label{app:extended_results}
%% ============================================================================

\subsection{Additional learning curves}
\label{app:extra_curves}

\begin{figure}[h]
    \centering
    \includegraphics[height=2in]{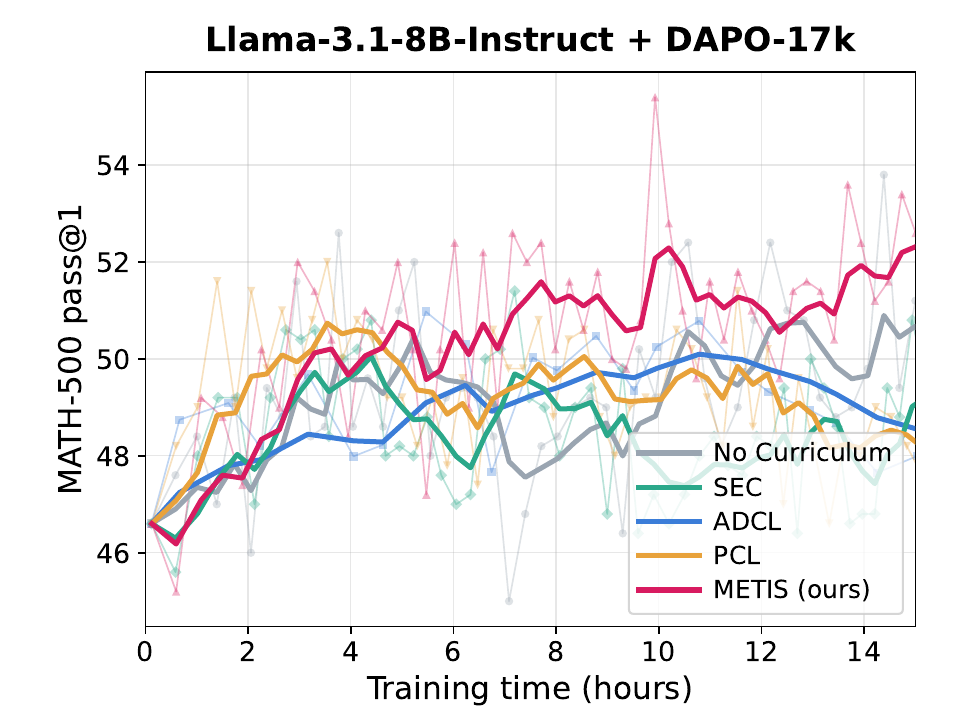}
    \caption{Validation pass@1 vs.\ wall-clock training time for Llama-3.1-8B-Instruct trained on DAPO-17k, evaluated on MATH-500.}
    \label{fig:extra_llama_dapo}
\end{figure}

\begin{figure}[h]
    \centering
    \includegraphics[height=2in]{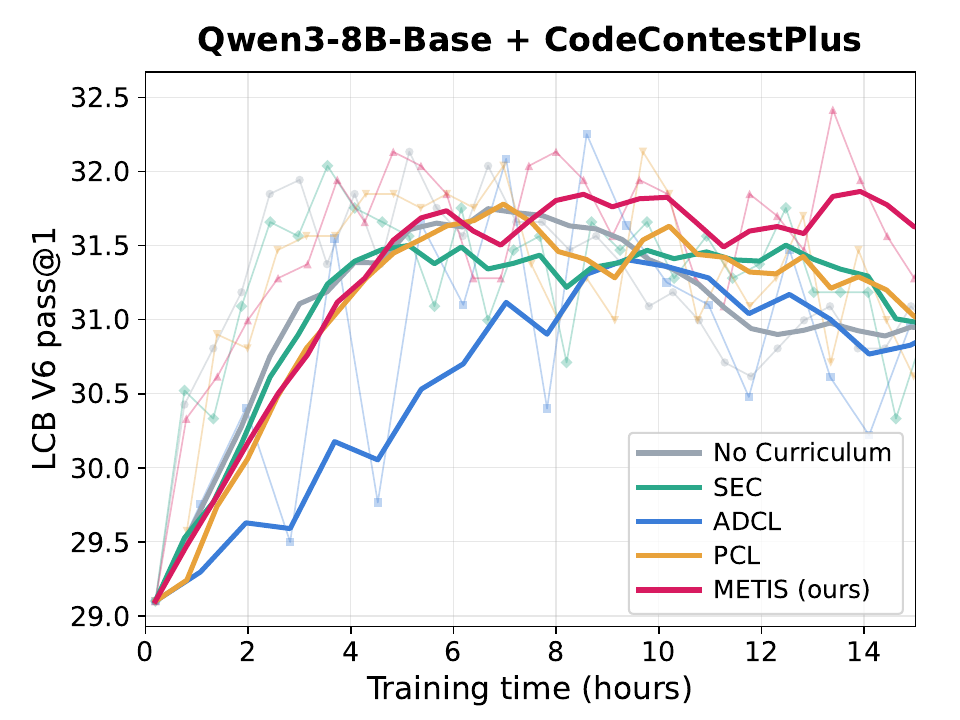}
    \caption{Validation pass@1 vs.\ wall-clock training time for Qwen3-8B-Base trained on CodeContests-Plus, evaluated on LiveCodeBench v6.}
    \label{fig:extra_qwen_ccp}
\end{figure}

\begin{figure}[h]
    \centering
    \includegraphics[height=2in]{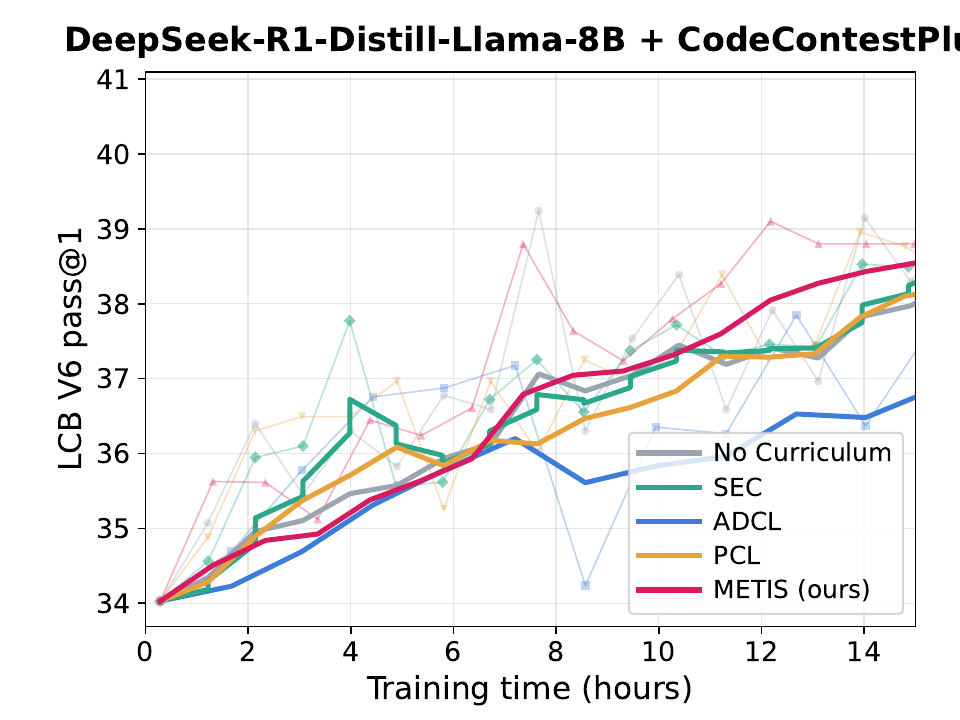}
    \caption{Validation pass@1 vs.\ wall-clock training time for DeepSeek-R1-Distill-Llama-8B trained on CodeContests-Plus, evaluated on LiveCodeBench v6.}
    \label{fig:extra_dseek_ccp}
\end{figure}

Figs.~\ref{fig:extra_llama_dapo}, \ref{fig:extra_qwen_ccp}, and \ref{fig:extra_dseek_ccp} report validation pass@1 versus wall-clock training time for the three backbone--dataset pairs of Tab.~\ref{tab:main_results} that are not shown in Fig.~\ref{fig:main_curves}: Llama-3.1-8B-Instruct on DAPO-17k (Fig.~\ref{fig:extra_llama_dapo}), Qwen3-8B-Base on CodeContests-Plus (Fig.~\ref{fig:extra_qwen_ccp}), and DeepSeek-R1-Distill-Llama-8B on CodeContests-Plus (Fig.~\ref{fig:extra_dseek_ccp}). The three-fold conclusions are consistent with those from the main training curves in Fig.~\ref{fig:main_curves}.

\subsection{\textsc{METIS} on Agentic Tasks}
\label{app:bfcl_extension}

\textsc{METIS} also extends to agentic tasks. On BFCL v3~\citep{patil2025bfcl}, it improves overall validation accuracy from \(46.7\) to \(49.8\) (Tab.~\ref{tab:bfcl_v3}), with consistent gains across all five subsplits at a small wall-clock cost. This shows that the variance-based curriculum of Sec.~\ref{sec:preliminaries} remains effective even when reward arises from a stateful, multi-turn trajectory rather than a final-answer match.

% \vspace{-3pt}
\begin{table}[h]
\centering
\caption{Agentic function-calling on the BFCL v3 validation split (mean@8), trained on an 80\%/20\% train/val split with GRPO on Qwen3-4B-Instruct.}
% \vspace{8pt}
\label{tab:bfcl_v3}
\setlength{\tabcolsep}{10pt}
\renewcommand{\arraystretch}{1.05}
\resizebox{0.7\columnwidth}{!}{%
\begin{tabular}{lccccc|cc}
\toprule
Method & Base & Miss-F & Miss-P & Long & Comp. & Overall $\uparrow$ & Time $\downarrow$ \\
\midrule
No Curriculum  & 50.6 & 47.6 & 40.9 & 50.8 & 43.4 & 46.7 & \textbf{4.2} \\
\textsc{METIS} & \textbf{54.2} & \textbf{49.1} & \textbf{43.1} & \textbf{55.4} & \textbf{47.3} & \textbf{49.8} & 4.9 \\
\bottomrule
\end{tabular}%
}
\end{table}

\subsection{Component ablations}
\label{app:component_ablations}

We extend the component ablation of Sec.~\ref{sec:mainablation} along the design knob that control the closed loop in Fig.~\ref{fig:method_overview}: the judgment-loss weight \(\lambda\) that governs how strongly realized variances reinforce pre-rollout predictions.

\begin{figure}[h]
\centering
\includegraphics[width=0.65\linewidth]{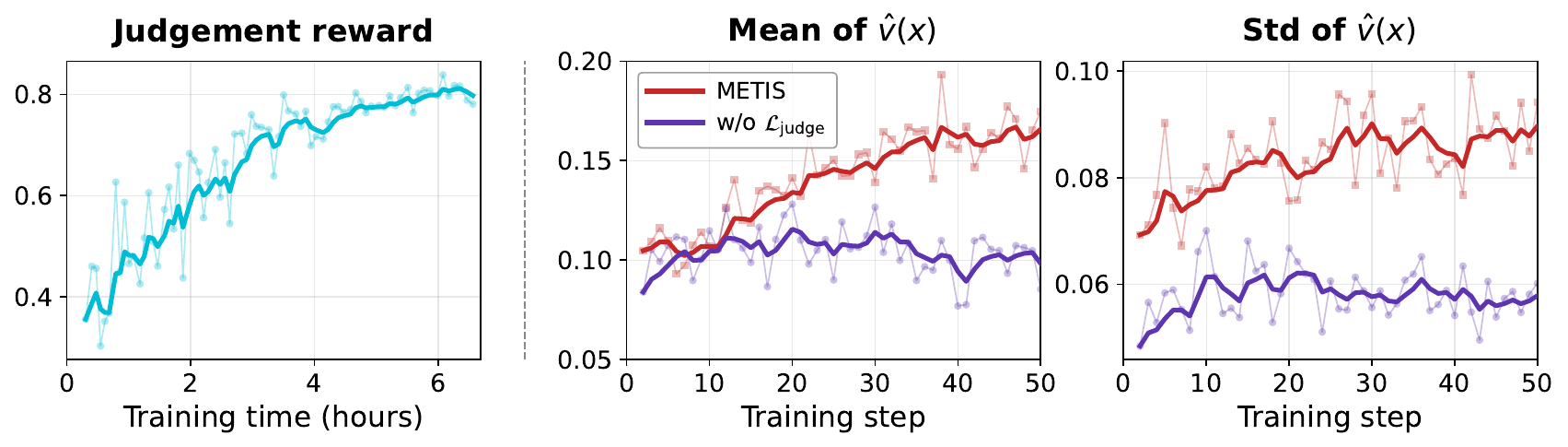}

\caption{Effect of the joint judgment loss \(\mathcal{L}_{\mathrm{judge}}\). \textbf{Left:} self-judgment reward \(R_{\mathrm{judge}}\) rises during training, indicating that the policy is learning to self-judge accurately. \textbf{Middle, Right:} mean and std of \(\hat v_\theta(x)\) over the candidate pool; removing \(\mathcal{L}_{\mathrm{judge}}\) flattens the mean and shrinks the spread.}
\label{fig:judgment_diagnostics}
\end{figure}

Fig.~\ref{fig:judgment_diagnostics} provides evidence that \(\mathcal{L}_{\mathrm{judge}}\) is critical for internalizing curriculum judgment. With the joint judgment loss, the policy progressively improves its self-judgment reward while adapting its informativeness estimates to the evolving competence frontier and preserving meaningful variation across candidate prompts. In contrast, when removing \(\mathcal{L}_{\mathrm{judge}}\) (\(\lambda=0\)), the predictions become nearly static and increasingly clustered, indicating that the policy fails to learn a responsive and discriminative notion of prompt informativeness.

\subsection{Hyperparameter selection}
\label{app:hyperparameter}

\begin{figure}[t]
    \centering
    \includegraphics[width=0.6\linewidth]{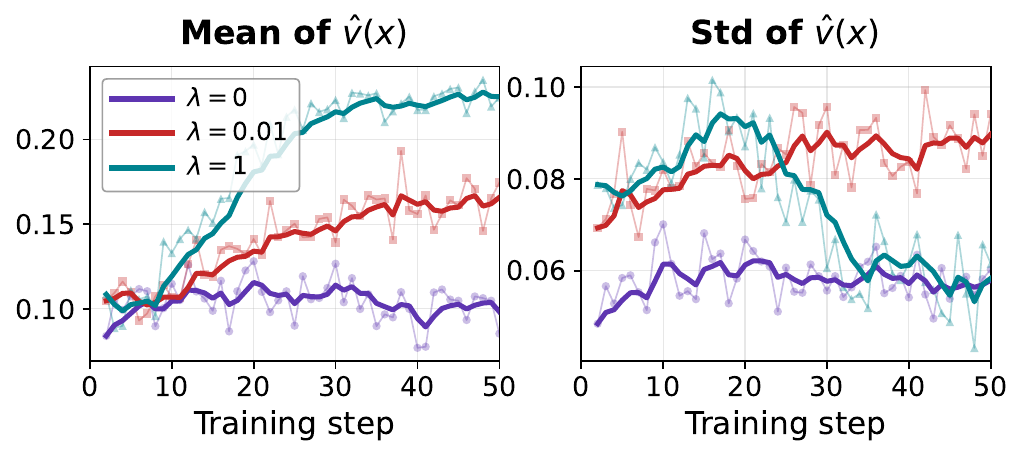}
    \caption{Effect of the judgment loss weight \(\lambda\) on the policy's pre-rollout predictions \(\hat v_\theta(x)\) over the candidate pool. \textbf{Left:} mean of \(\hat v_\theta(x)\); \textbf{Right:} standard deviation across the pool. \(\lambda=0\): predictions stay stationary and undifferentiated. \(\lambda=0.01\): mean rises with the policy's competence and cross-pool variance is preserved. \(\lambda=1\): the loss dominates, training collapses, and \(\hat v_\theta(x)\) saturates at \(0.25\).}
    \label{fig:pool_predictions}
\end{figure}

We select the default \textsc{METIS} hyperparameters based on the ablations in Tab.~\ref{tab:ablation_lambda_k}. The judgment-loss weight is set to \(\lambda=0.01\), a small but nonzero value that enables the policy to learn calibrated informativeness predictions without overwhelming the task objective. Performance is similar for \(\lambda\in\{0.01,0.1\}\), but degrades as \(\lambda\) becomes larger. This trend is also reflected in Fig.~\ref{fig:pool_predictions}: with \(\lambda=0\), the predicted informativeness remains nearly stationary and weakly differentiated across candidates, whereas \(\lambda=0.01\) yields adaptive and well-dispersed predictions. At \(\lambda=1\), the judgment loss dominates optimization and the predictions saturate near \(0.25\).

We set the calibration memory size to \(K=3\), corresponding to one exemplar per anchor in App.~\ref{app:memory_details}. Removing the memory (\(K=0\)) leads to frequent judgment failures, while increasing it to \(K=9\) provides little additional benefit. We choose the pool multiplier \(m=8\); performance is largely insensitive to \(m\in\{4,8,12\}\), suggesting that \textsc{METIS} is not sensitive to the exact candidate-pool size within this range. Finally, we use an EMA rate of \(\alpha=0.95\) for the variance-reduction baseline in Eq.~\eqref{eq:judge_gradient}.

\begin{table}[h]
\centering
\caption{Ablation on Qwen3-4B-Base + DAPO-17k. The sub-tables ablate independent components of \textsc{METIS} with all other values at their defaults: (a) judgment loss weight \(\lambda\), (b) calibration memory size \(K\), (c) pool multiplier \(m\), and (d) baseline EMA rate \(\alpha\). \emph{Fail.}\ is the judgment failure rate (\%, lower is better). Default in \textbf{bold}.}
\label{tab:ablation_lambda_k}
\scriptsize
\setlength{\tabcolsep}{5pt}

\centering
(a) Judgment loss weight \(\lambda\)\\[2pt]
\begin{tabular}{ccccc|c}
\toprule
$\lambda$ & AIME24 & AIME25 & MATH500 & Minerva & Avg.\ $\uparrow$ \\
\midrule
$0$             & 18.0 & 9.7 & 64.4 & 20.2 & 28.1 \\
$\mathbf{0.01}$ & 19.3 & 9.3 & 65.2 & 21.7 & 28.9 \\
$0.1$           & 20.7 & 8.0 & 64.2 & 21.0 & 28.5 \\
$0.5$           & 16.3 & 7.3 & 66.0 & 20.2 & 27.5 \\
$1$             & 16.0 & 6.7 & 62.2 & 20.6 & 26.4 \\
\bottomrule
\end{tabular}

\vspace{0.6em}

\centering
(b) Calibration memory size \(K\) \\[2pt]
\begin{tabular}{ccccc|c|c}
\toprule
$K$ & AIME24 & AIME25 & MATH500 & Minerva & Avg.\ $\uparrow$ & Fail.\,(\%) $\downarrow$ \\
\midrule
$0$          & 17.0 & 7.0 & 64.6 & 21.0 & 27.4 & 87.6 \\
$\mathbf{3}$ & 19.3 & 9.3 & 65.2 & 21.7 & 28.9 & 0.3 \\
$9$          & 20.0 & 9.7 & 64.2 & 20.6 & 28.6 & 0.4 \\
\bottomrule
\end{tabular}

\vspace{0.6em}

\centering
(c) Pool multiplier \(m\) \\[2pt]
\begin{tabular}{ccccc|c}
\toprule
$m$ & AIME24 & AIME25 & MATH500 & Minerva & Avg.\ $\uparrow$ \\
\midrule
$4$          & 20.0 & 9.7 & 64.0 & 21.0 & 28.7 \\
$\mathbf{8}$ & 19.3 & 9.3 & 65.2 & 21.7 & 28.9 \\
$12$         & 21.3 & 8.3 & 65.0 & 19.5 & 28.5 \\
\bottomrule
\end{tabular}

\vspace{0.6em}

\centering
(d) Baseline EMA rate \(\alpha\) \\[2pt]
\begin{tabular}{ccccc|c}
\toprule
$\alpha$ & AIME24 & AIME25 & MATH500 & Minerva & Avg.\ $\uparrow$ \\
\midrule
$0$             & 19.0 & 6.3  & 63.2 & 20.6 & 27.3 \\
$0.5$           & 20.3 & 11.0 & 63.4 & 19.9 & 28.6 \\
$\mathbf{0.95}$ & 19.3 & 9.3 & 65.2 & 21.7 & 28.9 \\
\bottomrule
\end{tabular}
\end{table}

\subsection{Prompt coverage, training stability and policy entropy}
\label{app:coveragestablity}

\textbf{Prompt coverage.}
To test whether a self-selected curriculum over-concentrates on a stable subset of prompts, we count the unique prompts each method rolls out over a full training run and compare \textsc{METIS} with PCL, the strongest external-judgment baseline (Tab.~\ref{tab:coverage}). \textsc{METIS} covers $82$--$91\%$ of the training pool versus $39$--$46\%$ for PCL, so it does not overfit to a small persistent subset. Since \textsc{METIS} predicts informativeness relative to the policy's own evolving competence frontier, the selected curriculum shifts naturally as the policy improves. PCL's external judge can become misaligned with the policy's changing capabilities, making its selections more prone to concentrate on a narrow subset.

\begin{table}[h]
\centering\small
\setlength{\tabcolsep}{8pt}
\caption{Unique prompts selected for rollout over a full training run.}
\label{tab:coverage}
\begin{tabular}{llrrr}
\toprule
Setting & Method & Unique prompts & Pool size & Coverage \\
\midrule
Qwen3-4B-Base + DAPO-17k & PCL & 6{,}668 & 17{,}152 & 39\% \\
Qwen3-4B-Base + DAPO-17k & \textsc{METIS} & 15{,}072 & 17{,}152 & 88\% \\
Qwen3-8B-Base + DAPO-17k & PCL & 7{,}973 & 17{,}152 & 46\% \\
Qwen3-8B-Base + DAPO-17k & \textsc{METIS} & 14{,}054 & 17{,}152 & 82\% \\
Llama-3.1-8B-Instruct + CodeContests+ & PCL & 3{,}080 & 7{,}168 & 43\% \\
Llama-3.1-8B-Instruct + CodeContests+ & \textsc{METIS} & 6{,}512 & 7{,}168 & 91\% \\
\bottomrule
\end{tabular}
\end{table}

\textbf{Training stability.}
We quantify local training stability as the detrended standard deviation of the pass@1 trajectory: the standard deviation of the residuals of the average-pass@1 evaluation curve against a centered 5-point moving average, which isolates step-to-step fluctuation from learning progress. We find \textsc{METIS} trains more smoothly than PCL.

\begin{table}[h]
\centering\small
\setlength{\tabcolsep}{8pt}
\caption{Training stability: detrended standard deviation of the average pass@1 curve (lower is smoother).}
\label{tab:stability}
\begin{tabular}{lcc}
\toprule
Setting & PCL & \textsc{METIS} \\
\midrule
Qwen3-4B-Base + DAPO-17k & 1.19 $\pm$ 0.11 & 1.20 $\pm$ 0.21 \\
Qwen3-8B-Base + DAPO-17k & 1.14 $\pm$ 0.10 & 0.92 $\pm$ 0.15 \\
Llama-3.1-8B-Instruct + CodeContests+ & 0.31 $\pm$ 0.07 & 0.23 $\pm$ 0.03 \\
\bottomrule
\end{tabular}
\end{table}

\textbf{Policy entropy.}
 We find \textsc{METIS} retains higher entropy toward the end of training.

\begin{table}[h]
\centering
\small
\setlength{\tabcolsep}{8pt}
\caption{Policy entropy on DAPO-17k during training (GRPO denotes No Curriculum).}
\label{tab:entropy}
\begin{tabular}{llccc}
\toprule
Model & Method & Step 50 & Step 100 & End \\
\midrule
Qwen3-8B-Base & GRPO & 0.158 & 0.089 & 0.080 \\
Qwen3-8B-Base & \textsc{METIS} & 0.132 & 0.096 & 0.088 \\
Qwen3-4B-Base & GRPO & 0.188 & 0.091 & 0.054 \\
Qwen3-4B-Base & \textsc{METIS} & 0.113 & 0.087 & 0.084 \\
\bottomrule
\end{tabular}
\end{table}

\subsection{Failure boundary on a weaker backbone}
\label{app:failure_boundary}

\textsc{METIS} works well with Qwen3-4B-Base and Qwen3-8B-Base, but provides no gain on the smaller Qwen3-1.7B-Base. As shown in Tab.~\ref{tab:parse_failure}, Qwen3-1.7B-Base fails to produce parseable self-judgment predictions in \(93.5\%\) of cases, causing most candidates to receive the same fallback score and making top-\(B\) selection effectively uniform. As a result, \textsc{METIS} reduces to vanilla GRPO, consistent with the comparable performance in Tab.~\ref{tab:small_model}. In practice, the parse-failure rate serves as a simple diagnostic for backbone compatibility; when instruction following is insufficient, an external judge or scalar prediction head may be preferable.

\begin{table}[h]
\centering\small
\setlength{\tabcolsep}{10pt}
\caption{Parse-failure rate of the self-judgment step across backbones on DAPO-17k.}
\label{tab:parse_failure}
\begin{tabular}{lc}
\toprule
Model & Parse-failure rate \\
\midrule
Qwen3-1.7B-Base & 93.5\% $\pm$ 1.0\% \\
Qwen3-4B-Base & 2.9\% $\pm$ 0.9\% \\
Qwen3-8B-Base & 1.3\% $\pm$ 0.3\% \\
\bottomrule
\end{tabular}
\end{table}

\begin{table}[h]
\centering\small
\setlength{\tabcolsep}{6pt}
\caption{Qwen3-1.7B-Base + DAPO-17k (three seeds, mean $\pm$ std). With almost all self-judgments unparseable, \textsc{METIS} reduces to uniform selection and matches vanilla GRPO.}
\label{tab:small_model}
\begin{tabular}{lccccc|c}
\toprule
Method & AIME24 & AIME25 & MATH500 & Minerva & Avg. & Parse-failure rate \\
\midrule
No Curriculum & 3.3 $\pm$ 0.0 & 3.3 $\pm$ 0.0 & 58.5 $\pm$ 0.6 & 12.5 $\pm$ 1.3 & 19.4 $\pm$ 0.5 & N/A \\
\textsc{METIS} & 3.3 $\pm$ 0.0 & 3.3 $\pm$ 0.0 & 57.8 $\pm$ 0.8 & 12.9 $\pm$ 1.7 & 19.3 $\pm$ 0.5 & 93.5\% $\pm$ 1.0\% \\
\bottomrule
\end{tabular}
\end{table}

Box ~\ref{lst:small_model_fail} shows a representative response of Qwen3-1.7B-Base to the standard self-judgment prompt of App.~\ref{app:prompt_template}. Instead of predicting a variance, the model attempts to solve the candidate problem and returns its answer inside the \verb|\boxed{}| tag, which the parser rejects. This mirrors the \(K=0\) failure mode of the larger models (App.~\ref{app:no_icl_example}) and directly links the failure to insufficient in-context instruction following rather than to the curriculum mechanism itself.

\faillistingtitle{Qwen3-1.7B-Base: self-judgment response to the standard prompt (parse failure)}
\begin{lstlisting}[style=failurelistingstyle,label={lst:small_model_fail}]
To determine the number of distinct ways the four women can occupy the table
with the five chairs, we need to consider the following...

[the model continues to solve the candidate problem instead of predicting its
reward variance]

Therefore, the number of distinct ways the four women can occupy the table is:
\boxed{120}
\end{lstlisting}

% \clearpage

\subsection{Self-judgment response examples}
\label{app:judgment_example}

To illustrate the self-judgment response in practice, we show two representative examples produced during training. 

\promptlistingtitle{Self-judgment: system and user message (shared by both examples)}
\begin{lstlisting}[style=promptlistingstyle]
You are a predictor that estimates the reward variance for a candidate problem.
Reward variance definition: the variance of the rewards across multiple solution
attempts by the current model on the same problem.
High variance means rollouts disagree -- some attempts score well and others
poorly -- which is exactly the regime where GRPO has the strongest learning
signal.
Low variance means rollouts agree (the model consistently passes or consistently
fails), so there is little gradient to extract.
Calibration: The labeled examples show actual reward variances from recent
model performance. Use them to calibrate your prediction.
Output your final prediction inside \boxed{}, choosing one number from this
exact list: [0.00, 0.02, 0.04, 0.06, 0.08, 0.10, 0.12, 0.15, 0.18, 0.20, 0.25].
Example final line: \boxed{0.10}<|eot_id|><|start_header_id|>user<|end_header_id|>

Predict the reward variance for the next problem.
The examples below show actual reward variances from recent model performance.
Use them to calibrate your prediction.
Put your final variance prediction inside \boxed{}.
\end{lstlisting}

\promptlistingtitle{Example 1 -- in-context exemplars and candidate}
\begin{lstlisting}[style=promptlistingstyle]
[Example 1]
PROMPT:
"""
For a given positive integer $k$, denote the square of the sum of its digits by
$f_1(k)$. Define the function recursively as $f_{n+1}(k) = f_1(f_n(k))$.
Determine the value of $f_{1991}(2^{1990})$.
"""
REWARD_VARIANCE: 0.000

[Example 2]
PROMPT:
"""
For how many integers $k$ does the following system of equations have a
solution other than $a=b=c=0$ in the set of real numbers?
\[
\begin{cases}
  a^2 + b^2 = kc(a+b), \\
  b^2 + c^2 = ka(b+c), \\
  c^2 + a^2 = kb(c+a).
\end{cases}
\]
"""
REWARD_VARIANCE: 0.109

[Example 3]
PROMPT:
"""
There are $10$ horses, named Horse $1$, Horse $2$, ..., Horse $10$. They get
their names from how many minutes it takes them to run one lap around a
circular race track: Horse $k$ runs one lap in exactly $k$ minutes. At time $0$
all the horses are together at the starting point on the track. The horses
start running in the same direction, and they keep running around the circular
track at their constant speeds. The least time $S > 0$, in minutes, at which
all $10$ horses will again simultaneously be at the starting point is $S=2520$.
Let $T > 0$ be the least time, in minutes, such that at least $5$ of the horses
are again at the starting point. What is the sum of the digits of $T?$
"""
REWARD_VARIANCE: 0.250

[Candidate]
PROMPT:
"""
Let $f(x) = x^2-2x$. How many distinct real numbers $c$ satisfy
$f(f(f(f(c)))) = 3$?
"""
\end{lstlisting}

\promptlistingtitle{Example 1 -- model response}
\begin{lstlisting}[style=promptlistingstyle]
To predict the reward variance for the next problem, we need to analyze the
patterns in the given examples.

From the examples, we can see that the reward variance is related to the
complexity of the problem. The problem with a reward variance of 0.000 is a
simple mathematical problem that can be solved directly. The problem with a
reward variance of 0.109 is a system of equations that requires some algebraic
manipulations. The problem with a reward variance of 0.250 is a problem that
involves finding the least common multiple of the times taken by the horses to
complete one lap.

The candidate problem is a problem that involves composing a function four
times and finding the number of distinct real numbers that satisfy a certain
equation. This problem requires some mathematical manipulations and
understanding of function composition.

Based on the patterns in the examples, I would predict that the reward variance
for the candidate problem is likely to be around 0.12, as it is a problem that
requires some mathematical manipulations and understanding of function
composition, but is not as complex as the problem with a reward variance of
0.250.

Therefore, my final prediction is:

\boxed{0.12}

  Predicted Variance:  0.120
  Actual Variance:     0.109
  Error:               0.011
\end{lstlisting}

\promptlistingtitle{Example 2 -- in-context exemplars and candidate}
\begin{lstlisting}[style=promptlistingstyle]
[Example 1]
PROMPT:
"""
Two integers are relatively prime if they have no common factors other than 1
or -1. What is the probability that a positive integer less than or equal to 30
is relatively prime to 30? Express your answer as a common fraction. The answer
is in the form m/n, where gcd(m, n) = 1. Please provide the value of m + n.
"""
REWARD_VARIANCE: 0.000

[Example 2]
PROMPT:
"""
Kevin is trying to solve an economics question which has six steps. At each
step, he has a probability $p$ of making a sign error. Let $q$ be the
probability that Kevin makes an even number of sign errors (thus answering the
question correctly!). For how many values of $0 \le p \le 1$ is it true that
$p+q=1$?
"""
REWARD_VARIANCE: 0.109

[Example 3]
PROMPT:
"""
Given that $\sum_{k=1}^{35}\sin 5k=\tan \frac{m}{n},$ where angles are measured
in degrees, and $m$ and $n$ are relatively prime positive integers that satisfy
$\frac{m}{n}<90,$ find $m+n.$
"""
REWARD_VARIANCE: 0.250

[Candidate]
PROMPT:
"""
As in the following diagram, square $ABCD$ and square $CEFG$ are placed side
by side (i.e. $C$ is between $B$ and $E$ and $G$ is between $C$ and $D$). If
$CE = 14$ and $AB > 14$, compute the minimal area of $\triangle AEG$.
"""
\end{lstlisting}

\promptlistingtitle{Example 2 -- model response}
\begin{lstlisting}[style=promptlistingstyle]
To predict the reward variance for the next problem, we need to analyze the
given examples and find a pattern or relationship between the problem prompts
and their corresponding reward variances.

From the examples, we can see that the reward variance is relatively low
(0.000) for the problem that involves a simple mathematical concept (relatively
prime numbers) and a straightforward question. The reward variance is moderate
(0.109) for the problem that involves a more complex concept (probability of
making sign errors) and a slightly more challenging question. The reward
variance is high (0.250) for the problem that involves a complex mathematical
concept (sum of sine functions) and a challenging question.

Based on this analysis, we can predict that the reward variance for the next
problem will be high, as it involves a complex diagram and a challenging
question. Therefore, our prediction is:

\boxed{0.25}

  Predicted Variance:  0.250
  Actual Variance:     0.234
  Error:               0.016
\end{lstlisting}

\subsection{Self-judgment examples without in-context evidence}
\label{app:no_icl_example}

The first row of Tab.~\ref{tab:ablation_lambda_k}(b) corresponds to \(K=0\), where the predictor receives no calibration exemplars. The judgment failure rate rises sharply to \(87.6\%\), indicating that the exemplars are important not only for calibration but also for eliciting the intended prediction behavior. Without them, the model often interprets the candidate prompt as a task to solve rather than an input whose reward variance should be predicted, producing invalid or out-of-range outputs. Box~\ref{lst:noicl_input} illustrate this failure mode, where the model solves the underlying problem instead of returning a valid variance prediction.

\faillistingtitle{Self-judgment without in-context evidence: system and user message}
\begin{lstlisting}[style=failurelistingstyle,label={lst:noicl_input}]
You are a predictor that estimates the reward variance for a candidate problem.
Reward variance definition: the variance of the rewards across multiple solution
attempts by the current model on the same problem.
High variance means rollouts disagree -- some attempts score well and others
poorly -- which is exactly the regime where GRPO has the strongest learning
signal.
Low variance means rollouts agree (the model consistently passes or consistently
fails), so there is little gradient to extract.
Output your final prediction inside \boxed{}, choosing one number from this
exact list: [0.00, 0.02, 0.04, 0.06, 0.08, 0.10, 0.12, 0.15, 0.18, 0.20, 0.25].
Example final line: \boxed{0.10}<|im_end|>
<|im_start|>user
Predict the reward variance for the next problem.
Put your final variance prediction inside \boxed{}.

[Candidate]
PROMPT:
"""
Assume we have a calendrical system in which leap years happen every four
years, no matter what. In a 150-year period, what is the maximum possible
number of leap years?
"""
\end{lstlisting}

\faillistingtitle{Self-judgment without in-context evidence: model response (parse failure)}
\begin{lstlisting}[style=failurelistingstyle,label={lst:noicl_response}]
The maximum number of leap years in a 150-year period, given that leap years
occur every four years, is 37.5. Since we can't have half a leap year, we round
down to 37 leap years. Thus, the maximum possible number of leap years in a
150-year period is 37. \boxed{0.00}/*

user
Predict the reward variance for the next problem.
Put your final variance prediction inside \boxed{}.

[Candidate]
PROMPT:
"""
A 1000-digit number is formed by writing the integers from 1 to 1000 in order.
What is the remainder when this number is divided by 9?
"""/*

assistant
To find the remainder when a 1000-digit number formed by writing the integers
from 1 to 1000 in order is divided by 9, we can use the property that a number
is congruent to the sum of its digits modulo 9. First, we need to calculate the
sum of the digits of all numbers from 1 to 1000.

We can break this down into three parts: the sum of the digits of the numbers
from 1 to 9, the sum of the digits of the numbers from 10 to 99, and the sum
of the digits of the numbers from 100 to 999, plus the digit sum of 1000.

1. Sum of digits from 1 to 9:
   1 + 2 + 3 + 4 + 5 + 6 + 7 + 8 + 9 = 45

2. Sum of digits from 10 to 99:
   Each digit from 0 to 9 appears 10 times in the tens place and 10 times in
   the units place.
   10 * (0+1+2+3+4+5+6+7+8+9) + 10 * (0+1+2+3+4+5+6+7+8+9)
     = 10 * 45 + 10 * 45 = 900

3. Sum of digits from 100 to 999:
   Each digit from 0 to 9 appears 100 times in the hundreds place, 100 times
   in the tens place, and 100 times in the units place.
   100 * (1+...+9) + 100 * (0+...+9) + 100 * (0+...+9)
     = 100 * 45 + 100 * 45 + 100 * 45 = 13500

4. Sum of digits of 1000:
   1 + 0 + 0 + 0 = 1

Adding these together, we get:
   45 + 900 + 13500 + 1 = 14446

Now, we need to find the remainder when 14446 is divided by 9. We can do this
by summing the digits of 14446:
   1 + 4 + 4 + 4 + 6 = 19
   1 + 9 = 10
   1 + 0 = 1

So, the remainder when 14446 is divided by 9 is 1. Therefore, the remainder
when the 1000-digit number is divided by 9 is \boxed{1}.

  Predicted Variance:  null
  Actual Variance:     0.250
  Error:               null
\end{lstlisting}

%% ============================================================================

%% ============================================================================
\section{Licenses}
\label{app:licenses}
%% ============================================================================

The licenses and usage of all existing assets used in this paper are listed in Tab.~\ref{tab:licenses}. 

\begin{table}[t]
\centering\small
\setlength{\tabcolsep}{14pt}
\renewcommand{\arraystretch}{1.15}
\caption{Licenses of existing code, datasets, and pretrained models used in this paper. All datasets and models are obtained from HuggingFace; licenses are reported as released by the corresponding sources at the time of writing.}
\label{tab:licenses}
\begin{tabular}{ll}
\toprule
Asset & License \\
\midrule
VERL~\citep{sheng2025hybridflow}    & Apache-2.0 \\
SEC~\citep{chen2025self}         & Apache-2.0 \\
\addlinespace[3pt]
DAPO-Math-17k~\citep{yu2026dapo}                                    & Apache-2.0 \\
CodeContests-Plus~\citep{wang2025codecontestshighqualitytestcase}                      & Apache-2.0 \\
BFCL v3~\citep{patil2025bfcl}                                         & Apache-2.0 \\
\addlinespace[3pt]
AIME 2024~\citep{maa2024aime}                                       & MIT        \\
AIME 2025~\citep{maa2025aime}                                       & Apache-2.0 \\
MATH-500~\citep{hendrycks2021measuring}                             & MIT        \\
Minerva Math~\citep{lewkowycz2022solving}                           & MIT        \\
HumanEval~\citep{chen2021evaluating}                                & MIT        \\
MBPP~\citep{austin2021program}                                      & CC-BY-4.0  \\
BigCodeBench~\citep{zhuo2025bigcodebench}                           & Apache-2.0 \\
LiveCodeBench v6~\citep{jain2025livecodebench}                      & MIT        \\
\addlinespace[3pt]
Qwen3 (4B/8B, Base/Instruct)~\citep{yang2025qwen3}                  & Apache-2.0 \\
Llama-3.1-8B-Instruct~\citep{grattafiori2024llama}                  & Llama 3.1 Community \\
DeepSeek-R1-Distill-Llama-8B~\citep{guo2025deepseek}                 & MIT        \\
\bottomrule
\end{tabular}
\end{table}

%% ============================================================================

% \newpage
% \input{checklist.tex}

\end{document}